\documentclass[a4paper,9pt]{extarticle}

\usepackage[utf8]{inputenc}
\usepackage[T1]{fontenc}

\usepackage[a4paper, margin=2.5cm]{geometry}

\usepackage{lmodern}
\usepackage{comment}
\usepackage{listings}
\usepackage{tikz}
\usepackage{graphicx}
\usepackage{float}
\usepackage{authblk}
\usepackage{multirow}
\usepackage{subfig}

\usepackage[backend=biber,
citestyle=numeric,
style=numeric-comp,
sortcites=true,
natbib=true,
url=false,
doi=false,
isbn=false,
sorting=nty,
]{biblatex}
\title{A simulation framework for autonomous lunar construction work}

\author[1]{Mattias Linde}
\author[1]{Daniel Lindmark}
\author[1]{Sandra Ålstig}
\author[1,2]{Martin Servin$^{*}$}

\affil[1]{\footnotesize Algoryx Simulation AB, Umeå, Sweden}
\affil[2]{\footnotesize Department of Physics, Umeå University, Umeå, Sweden}
\affil[*]{\footnotesize Corresponding author: martin.servin@umu.se}

\date{}

\begin{document}

\maketitle

\begin{abstract}
We present a simulation framework for lunar construction work involving multiple autonomous machines. 
The framework supports modelling of construction scenarios and autonomy solutions, execution of the scenarios 
in simulation, and analysis of work time and energy consumption throughout the construction project.
The simulations are based on physics-based models for contacting multibody dynamics 
and deformable terrain, including vehicle-soil interaction forces and soil flow in real time. 
A behaviour tree manages the operational logic and error handling, which enables the 
representation of complex behaviours through a discrete set of simpler tasks in 
a modular hierarchical structure. High-level decision-making is separated 
from lower-level control algorithms, with the two connected via ROS2. 
Excavation movements are controlled through inverse kinematics and tracking 
controllers. The framework is tested and demonstrated on two different 
lunar construction scenarios that involve an excavator and dump truck with 
actively controlled articulated crawlers.
\end{abstract}

\subsubsection*{Keywords}
Lunar construction; Excavation; Physics-based simulation; Automation; Behaviour trees.

\section{Introduction}
Planning construction work on the Moon, such as the Artemis project lunar base, is a
challenging task. Many aspects remain unknown, including local ground conditions, how the
machines should be designed and operated to accomplish the task within the stipulated
time and adhere to limited energy budgets. The machines should possess a high degree of
autonomy due to the significant delays associated with remote control from Earth.
Additionally, it is important that multiple machines can work in a well-coordinated manner, 
avoiding idle times and utilizing the equipment as efficiently as the energy supply allows for. 

To succeed in this and similar tasks, simulation-based tools are crucial for studying how 
different configurations of the lunar conditions, machine design, and control system affect 
the time and energy consumption in a given construction project.
These tools are also essential for detecting weaknesses in the entire approach and reducing
the risk of severe failures. Additionally, such tools can also be used for generating large
amounts of synthetic data for training deep learning models. 
To effectively iterate over numerous scenarios, each involving long sequences of 
machine operations, the simulations must be highly computationally 
efficient while remaining sufficiently realistic in terms of forces, 
power consumption, material flow, and vehicle behaviour.
Furthermore, it must be easy to define and test different 
work plans, autonomous functions, and combinations of sensors.

To this end, we develop a framework for simulating autonomous ground construction
tasks in a lunar environment. The framework leverages physics-based simulations of autonomously
controlled construction machines interacting with dynamic terrain, utilizing the AGX Dynamics
realtime physics engine \cite{AGX2025,servin2021multiscale}. 
The systems' autonomous control is modeled using behaviour trees 
\cite{colledanchise2018behaviourtrees,iovino2022behaviortrees},
which enables the representation of complex behaviours (tasks) through a discrete set of
simpler tasks organized in a hierarchical and modular structure.
High-level decision making is performed in the behaviour tree, whereas low-level
control algorithms are handled by skills on the different machines.
This approach makes it easy to replace and introduce new capabilities and
new machines. Communication between the machine's controller and the current 
state of the lunar environment occurs via ROS2, simplifying
the replacement of virtual components with their physical counterparts.

Our main contribution in the present work is a description of said framework and 
results from testing it on two scenarios, namely extracting 
lunar regolith and excavating a predefined ground structure. The tests
involve one excavator and one dump truck, but the framework is inherently 
designed to support many different machines. The work illustrates that
the framework supports analysis
of how the construction performance and energy use depend on different 
environmental factors.

\subsection{Related work}
There are few prior publications on simulation frameworks for autonomous lunar construction.
A recent work is the framework introduced in \cite{batagoda2024physics}. It supports synthetic sensors,
vehicle dynamics, and terramechanics using the Chrono physics engine. The terramechanics models
include the Soil Contact Model (SCM) for wheel-terrain interaction, and continuum (SPH) and
discrete (DEM) soil models for general use, including excavation and bulldozing. The
SynChrono module give support for multi-agent systems, e.g., sharing of state information with
synchronised but distributed simulation across computing nodes. The simulations involving
ground construction work are too computationally intense for use in applications with 
realtime or higher performance. Case studies in the paper focus on generation and analysis
of synthetic images using physically-based rendering for realistically capturing lunar lighting
conditions.

In \cite{thangavelautham2017autonomous}, autonomous multirobot excavation for lunar applications
was controlled using artificial neural tissue (ANT), trained using multi-agent grid world
simulations and evaluated using a higher-fidelity 3D simulation based on the tool Digital
Spaces with the fundamental earthmoving equation for bucket-soil interaction
\cite{el2008infrastructure}. Modeling or analysis of power transmission and consumption is not
addressed. In simulation, the ANT controller was found to achieve 30\% better fitness 
(target grid cells being excavated) than with a conventional neural network.

From NASA, the Digital Lunar Exploration Sites Unreal Simulation Tool (DUST)
\cite{bingham2023} makes lunar terrain data and several computational tools available in the
Unreal Engine 3D environment to support the Artemis mission. NASA-Ames and Open Robotics
have developed a Gazebo-based lunar rover simulator. Wheel-terrain interaction is
accounted for by empirical drawbar-pull and slip models, and terrain deformations are not supported.

OmniLRS is a simulation framework under development that uses IsaacSim from NVIDIA for
photorealistic simulation of planetary rovers on lunar terrain \cite{richard2024omnilrs} with a
data-driven wheel-terrain interaction model \cite{kamohara2024modeling}. As far as we are aware, 
the framework does not support excavation or bulldozing.

In \cite{sutoh2024development}, a tracked excavation rover platform was studied in
simulations based on the Vortex physics engine from CM Labs and by field experiments.
No conclusions were reported.

Teleoperation tasks for lunar surface construction, in particular excavation,
were studied in virtual reality using the Unity Engine in \cite{seo2024exploratory} 
to examine the effect of time delay. The outcomes indicate that time delays 
significantly degrade task performance, and the operators modify their control strategies 
to cope with the time delay conditions.

None of the related work offers the combined functionality of realtime physics-based 
simulation of lunar construction work and representation of complex behaviour 
of autonomous machines for executing the construction work plan.

\section{Method}
This section introduces the underlying methods for physics-based simulation,
behaviour tree-based task planning for multiple mobile machines, and ROS
for interconnecting them.

\subsection{Physics-based simulation}




For modeling and simulation, we rely on the physics engine AGX Dynamics \cite{AGX2025}.
It is based on the framework for contacting multibody dynamics introduced in
\cite{Lacoursiere2007}, extended to the discrete elements in \cite{Servin2014}, and to dynamic
deformable terrain in \cite{servin2021multiscale}. Specifically, it uses maximal coordinate representation
in terms of rigid bodies and kinematic constraints for joints, motors, and frictional contacts.
This has been used in previous work to develop machine learning models for autonomous
control of earthmoving equipment \cite{backman2021continuous,aoshima2023predictor} and 
active suspension system for rough terrain navigation \cite{wiberg2021control} with successful transfer of 
controllers trained in simulation to reality of working \cite{wiberg2024sim}.

The time-stepper and solver are optimized particularly for fixed timestep
realtime simulation of multibody systems with non-ideal constraints and non-smooth dynamics \cite{lacoursiere2011}. 
The hybrid direct-iterative solver supports fast and stable simulations at 
high accuracy for mathematically stiff and ill-conditioned systems \cite{lacoursiere2011},
such as robots and vehicles, 
and scalability to large-scale dynamic contact networks, 
such as for granular systems, at the price of lower accuracy \cite{Servin2014}.

For deformable terrain, AGX uses the multiscale model described in
\cite{servin2021multiscale}. It has been demonstrated to produce digging forces 
and soil displacements with an accuracy of 75-90\% of that of a resolved discrete 
element method (DEM), coupled with multibody dynamics, and with field tests involving 
full-scale construction equipment \cite{aoshima2024examining}.
The model can be regarded both as a reduced-order model of DEM and as a multibody
dynamics generalization of the FEE. The idea is to dynamically resolve only the part
of the terrain inside a well-localized region of shear failure. The terrain is 
assigned a set of bulk mechanical parameters for its physical behaviour in nominal
bank state. When a digging tool contacts the terrain surface, a zone of active
soil displacement is predicted. Only inside the active zone is the soil resolved into
particles. These are modeled using DEM with specific mass density and contact
parameters that ensure a bulk mechanical behaviour consistent with the assigned bulk
parameters through a pre-calibration procedure. See, Ref.~\cite{servin2021multiscale} for details.
The reaction force on the bucket consists of frictional-cohesive contacts 
with an aggregate body, which inherits the physical properties of the soil in the active 
zone, and a penetration resistance constraint, which is a function of the bucket geometry, 
soil parameters, and the soil pressure at the cutting edge.

The particle size varies dynamically in the multiscale model by mass exchange
between particles and with the terrain. This provides a balanced tradeoff
between fine resolution, close to the bucket cutting edge and the terrain surface, and 
coarser elsewhere. To avoid the DEM simulation being a computational bottleneck, 
large particles are favored by the algorithm.
With the multiscale approach, the errors in digging forces remain smaller than if the
terrain were composed of DEM particles. However, large particles misrepresent the real
soil flow and distribution inside the bucket. Specifically, the volume fraction
is overestimated at the soil-bucket interface. Consequently, with large particles 
the bucket filling is typically underestimated unless one compensates by assigning
the particles a higher mass density by a factor proportional to the particle cross-section
area over the bucket's internal surface area.

The terrain model can also be assigned to dynamic bodies, such as the load-carrying 
truckbed of a dump truck. Terrain particles that are dumped and settling to rest, are 
converted to a local terrain that co-moves with the dump truckbed and adds to its
mass and inertia tensor.  When the truckbed is raised and tilted to a critical angle, the 
the terrain starts avalanching and is resolved into particles until they come to rest 
again.  


\subsection{Behaviour trees}
\newcommand{\SUCCESS}{\textit{Success}}
\newcommand{\FAILURE}{\textit{Failure}}
\newcommand{\RUNNING}{\textit{Running}}
To achieve modular, scalable, and explainable machine behaviours, behaviour trees (BT) are utilized.
They are widely used in artificial intelligence applications for representing task switching policies.
See \cite{iovino2022behaviortrees} and \cite{colledanchise2018behaviourtrees} for an in-depth treatment.
A BT is a rooted tree structure where the nodes are connected with parent-child relationship. When the
tree is \textit{executed}, or \textit{ticked}, traversal begins at the root and proceeds according to
the types of nodes and the status of the nodes.
Each node returns one of the following status values, \SUCCESS{}, \FAILURE{}, \RUNNING{} and
the status values affect the traversal for the common node types in the following way:

\begin{description}
	\item[Sequence] A sequence node ticks all its children in order. If any child returns \FAILURE{} or \RUNNING{}, the sequence halts and returns the same
		status. The sequence returns \SUCCESS{} only if all children succeed.
	\item[Selector]  A selector node, also called fallback node, ticks its children in order until one returns \RUNNING{} or \SUCCESS{} and the selector will
		return the same status. If all children fail, then the selector also returns \FAILURE.
	\item[Parallel]  A parallel node ticks all children concurrently. A policy (e.g. succeed on one, succeed on all, succeed on specific child)
		determines how the children's statuses are aggregated into the parallel nodes return status.
	\item[Decorator] A decorator is a node with a single child that can change the child status, e.g. an Inverter swaps \SUCCESS{} and \FAILURE{}.
	\item[Condition] A condition is a leaf node that checks a condition and returns either \SUCCESS{} or \FAILURE{}.
	\item[Task]      A task or action is a leaf node that represents an action. It returns \RUNNING{} while the action is ongoing, \SUCCESS{} when the
		action is completed, and \FAILURE{} if the action could not be performed.
\end{description}

\subsection{Robot Operating System}
Robot Operating System (ROS) is a middleware framework for robot software development \cite{quigley2009ros}. It provides tools, libraries, and conventions for creating complex and robust robot behaviour across different robotic platforms. With standardised interfaces, ROS encourages modular development with packages for specific robotic functionalities and services designed for heterogeneous computer clusters, including hardware abstraction and low-level device control.
With ROS2, it also functions as a middleware for communication \cite{Macenski:ROS2:2022}. The message-passing interface enables independent processes, nodes, to communicate via publications and subscriptions.

\section{Simulation framework}
This section begins with a general overview of the autonomous
construction framework, followed by a detailed explanation of how its
components are used to control machines for performing autonomous tasks.
\begin{figure}[H]
    \centering
    \includegraphics[width=0.75\textwidth,]{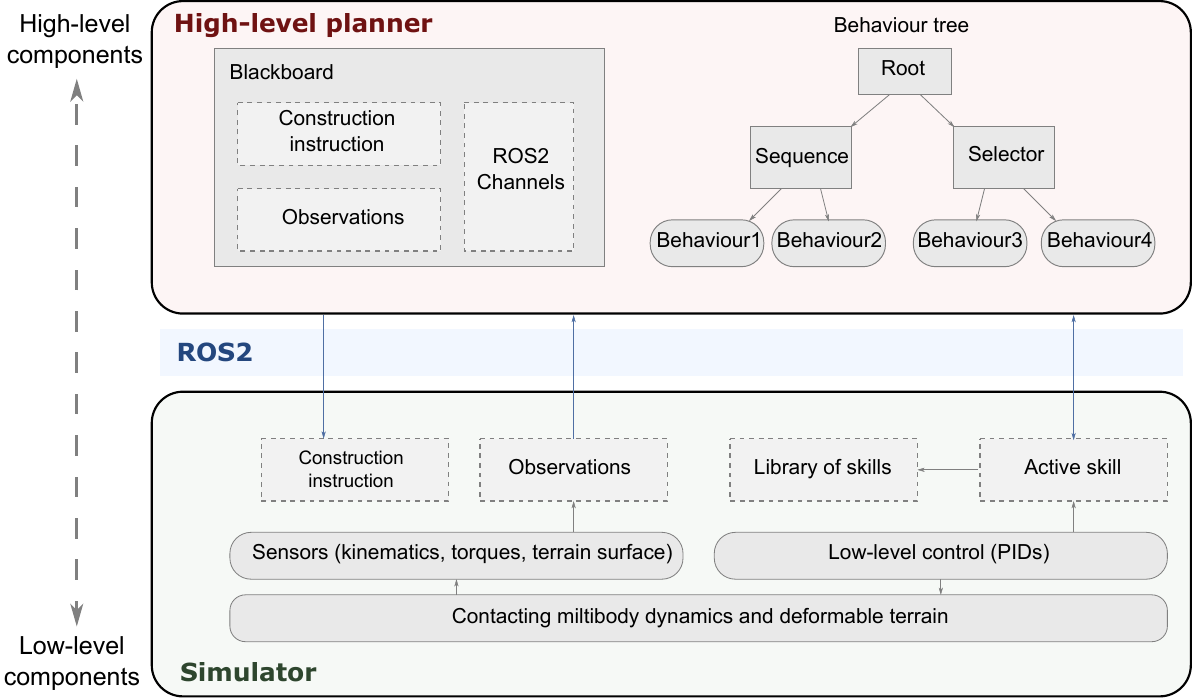}
    \caption{Overview of the framework with its two main components, the
	high-level planner and the simulator interconnected with ROS.}
    \label{fig:framework-overview}
\end{figure}

The autonomous simulation framework is implemented in Python and consists of two main
components: \emph{a high-level planner} and a \emph{simulator}. These components
run as separate processes and communicate using ROS2.
This separation introduces a key feature: the planner does not have direct access
to the ground truth of the world state, as that information resides within the simulator.
Instead, the planner must rely on observations being sent from the machines to
update its own world state. Although there is no technical barrier preventing
the components from being in the same process, maintaining them as separate processes offers
several advantages. It enhances modularity, facilitates the replacement of individual parts and supports
the development of more robust autonomous systems where perfect information cannot be assumed.
Furthermore, this architecture simplifies future extensions, such as
introduction of extra communication delays or disturbances.  An overview
illustration of the framework is found in Fig.~\ref{fig:framework-overview}.

\subsection{High-level planner}
The tasks to be performed by the planner are defined using a behaviour tree.
This tree is composed of PyTrees' built-in \emph{composites} and \emph{decorators},
along with a set of predefined framework behaviours, and when necessary,
user-defined behaviours. Tree nodes can share data via a common blackboard and may
add information for future node ticks. The framework makes use of the \emph{memory}
feature for sequence nodes --- where the status \emph{running} results in the sequence being
restarted from the running node on the next tick instead of from the beginning.

\begin{figure}[ht]
	\centering

  \includegraphics{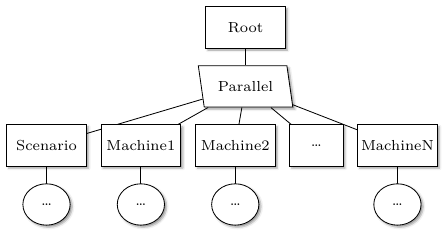}

	\caption{Example structure on a high level for a behaviour tree with multiple machines.}
\label{fig:example-tree}
\end{figure}

Users define the behaviour tree using a structured text format. Each non-empty line
includes an \emph{optional node name}, a \emph{node specification} and an
\emph{optional machine context}. Indentation levels are used to indicate parent-child
relationship between nodes. Comments can also be added using a simple annotation format.
The node specification can either reference a class name or a Python method thay returns
a node or subtree. This allows for modular reuse of subtrees across different machines,
minimizing code duplication. The machine context, specified as \texttt{-> machinename},
indicates which machine a subtree should interact with. This enables the reuse of identical
node types multiple times within a tree, each instance operating on a different machine.
An example of textual representation of a behaviour tree is illustrated in Fig.~\ref{fig:example-file} with the
corresponding tree representation in Fig.~\ref{fig:example-tree}. 

\lstdefinestyle{customstyle}{
  basicstyle=\ttfamily\small,
	commentstyle=\color{gray},
  morecomment=[l]{\#},
	breaklines=true,
	tabsize=2,
}
\lstset{style=customstyle}

\begin{figure}[ht]
%
%
	\begin{lstlisting}
Root: Sequence
  	Parallel: ParallelSuccessOnFirst
  		# Scenario completed? -> Tree Success
			Scenario: Sequence
        ...
			# Decorator between machine subtrees
			# and parallel node are omitted.
	  	Machine1: Sequence -> excavator1
	    	...
  		Machine2: Sequence -> dumptruck1
	    	...
\end{lstlisting}
\caption{Example of a textual representation of a behaviour tree. Dots are used to indicate 
omitted nodes or subtrees to keep the example short.}
\label{fig:example-file}
\end{figure}

\subsection{Simulator}
The simulator component features a dynamic terrain and one or more machines.
Each machine is equipped with a set of \emph{skills} - capabilities that can be activated by the planner.
Skills may also accept optional input data required for their execution.
These skills correspond to specific actions that machines can perform through their low-level control systems.

Communication between the planner and simulator occurs over multiple ROS2 topics.
Each node in the behaviour tree can have an associated skill on the machine, and the
communication topic incorporates the machine context.
For example, a topic could use \texttt{/\{machine context\}/target/drive} as name template.
This naming convention allows generic nodes within the planners behaviour tree to interact with
specific machines, supporting modular and scalable control logic.
The same machine-context naming scheme is used for the planner to subscribe to telemetry and evaluation data
from machines. These topics provide metrics such as execution time, energy consumption and skill status.
Collecting this data enabled the assessment of a work plan's efficiency and overall system performance.

\textbf{Case study:} \emph{Excavating terrain to match a target profile.}
To execute an excavation task using the autonomous framework, the process begins by selecting
which machines will participate in the operation. Afterwards, a logical plan must be formulated
outlining how the task should be executed, and then translated into a behaviour tree structure.
With the tree setup, the next step involves preparing an initial configuration, currently
done primarily using JSON files, which ensure that both the simulator and planner share a consistent
initial view of the scenario configuration. Once this setup is complete, both components can be launched,
allowing the machines to begin autonomous operation.

Assuming a similar structure as depicted in Fig.~\ref{fig:example-tree} and that
a single excavator and one dump truck are involved, the behaviour tree structure
offers a highly intuitive way to model and reason about the roles and actions of
each machine. The use of a \emph{parallel} node within the tree architecture allows
each machine's subtree to operate independently. This modularity simplifies the addition
of new tasks or machines without devolving into the complexity associated with
traditional state machines.

\begin{figure}[ht]
%
%
	\centering

  \includegraphics{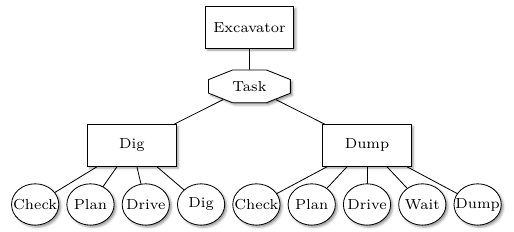}

	\caption{Subtree for an excavator.}
	\label{excavator-subtree}
\end{figure}

A logical breakdown of the excavator's tasks is: if the bucket is empty, plan where
digging should be performed and plan a drive trajectory towards the location, drive, and
dig. If the bucket is non-empty, then a dig cycle has been completed, and then a dump
cycle should be performed. If the bucket has material, plan where the material should be
offloaded and plan a drive trajectory to the location, drive, wait for a dump truck to
be in position, offload material. This workflow is illustrated in Fig.~\ref{excavator-subtree}.

The tasks for a dump truck can be divided and expressed in a similar way with two subtrees:
Plan where to accept material and plan a route to the loading location if the truckbed is not
full, then drive to the location. If the truckbed is full, plan where to offload material and
plan a route to the target, wait until no excavator is offloading material,
drive, and empty the truckbed.

\begin{figure}[ht]
%
%
	\centering

  \includegraphics{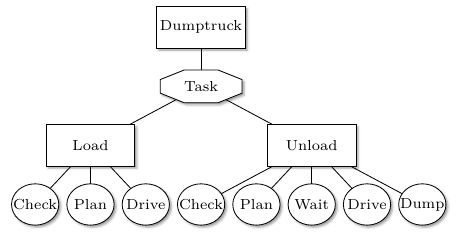}

	\caption{Subtree for a dump truck.}
	\label{dumptruck-subtree}
\end{figure}

The nodes shown in the subtrees in Fig.~\ref{excavator-subtree} and
\ref{dumptruck-subtree} map to functionality implemented in the framework.
Nodes responsible for various types of planning do not activate machine
skills directly. Instead, they carry out computation and store the result
on the blackboard, allowing subsequent nodes to make use of that information.
For instance, a drive node retrieves the precomputed route information, activates
a driving skill on a machine, and passes along waypoint data required for execution.

To maintain execution integrity, especially in scenarios where certain tasks may
fail and require re-execution, \emph{FailureIsRunning} decorators are placed between
a machine's subtree and the parallel node. These decorators prevent PyTrees from
reinitializing all subtrees in response to a failure, thereby supporting selective
re-planning and robust recovery mechanisms.

The final portion of the behaviour tree addresses scenario completion.
Determining when a scenario should be considered complete depends on how tasks,
primarily the digging, are organized and carried out.
A common strategy involves dividing the construction site into a local grid of cells,
with the current cell index stored on the blackboard.
During dig planning, the target profile is compared against the current world
state for the active grid cell. If the current terrain matches the
target within a predefined tolerance, the cell index is increased and planning
restarted. If material needs to be removed, a dig trajectory is planned
for the current active cell, taking the target profile into account.
The scenario is considered complete when the final grid cell has been processed.
This condition can be verified by a behaviour tree node that reads
from the blackboard and checks if all grid cells have been successfully handled.


\subsection{Motion planning and control}

When the excavator should perform a certain task, for example digging out a section of the
terrain, local joint-space values are needed that result in the end-effector performing the
desired motion in task space. To compute these joint values, the Inverse Kinematics (IK)
module in AGX Dynamics is used.
The excavator model in the simulation contains all the information that is needed to do
IK computations without specifying additional link lengths or joint positions. The attachment
frames for the constraints already hold all the needed information, including possible
joint ranges. Therefore, redesign of the machine can be done during prototyping without
maintaining redundant values that risk becoming out of sync. The types of constraints that
are supported are Hinge (rotational), Prismatic (sliding) and Lock (fixed).

The API for computing Inverse Kinematics in agxModel accepts an AffineMatrix4x4 that
contains the desired end-effector transform. That transform might not be reachable due to
being e.g outside of the working range, or the desired orientation might not be achievable
due to the degrees of freedom available in the mechanical structure.

To simplify the usage of IK computations, the LunarExcavator class has a method,
\texttt{calculate\_ik}, that accepts a world position and an angle versus the ground for the
digging direction. This is to avoid requesting rotations in directions the excavator bucket can
not physically rotate. Despite this, it is possible to request end-effector transforms which can
not be realised by the machine.

When a transform can not be reached with sufficient precision, an error code is returned as
well as the best joint values the IK-solver could find. These joint values can be used in a
forward computation to see where the end-effector could go. Comparing the reachable
transform with the requested transform is something a control algorithm can do as a step to
decide how to proceed.


\section{Numerical experiments}
The framework was tested for its ability to support the analysis of autonomy
solutions for lunar ground construction with multiple machines. 
This was done with numerical experiments covering two scenarios,
one where lunar material is loaded and transported from an excavation area to
a dump site, and a second one where the goal is to create a ground construction
of a given shape. The simulations run approximately in realtime with a time-step 
of 10 ms with the behaviour tree accounting for about 0.3 ms of the computing 
time per time-step. Simulations were run on an AMD Ryzen 9 5900X desktop processor
with 12 CPU cores and 3.7 GHz clock speed (GPU does not accelerate these simulations). 
Video material from the simulations can be found at \url{https://www.algoryx.se/papers/lunar-construction}.

\subsection{Construction machines and lunar environment}
Both scenarios involve an excavator and a dumper. They are both
equipped with crawlers with actively controlled sub-crawlers that promote
high terrain mobility. When turning, the sub-crawlers are lifted to reduce
turning resistance. While excavating and driving straight, the sub-crawlers are
lowered to reduce sinkage and actively controlled to smoothly follow the
terrain surface. The speed of the crawlers are controlled individually via
a rotational motor for each sprocket. The motion of the sub-crawlers relative
to the main crawler is also controlled via rotational motors.

\begin{figure}[H]
    \centering
    \includegraphics[width=0.4\textwidth]{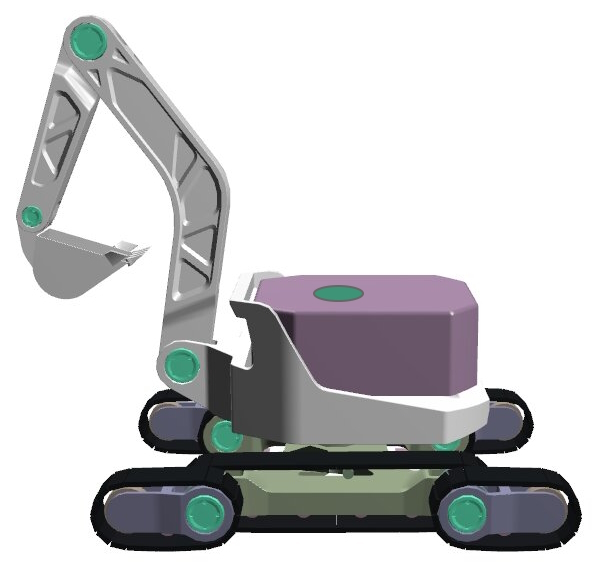}
    \hspace{3mm}
    \includegraphics[width=0.4\textwidth]{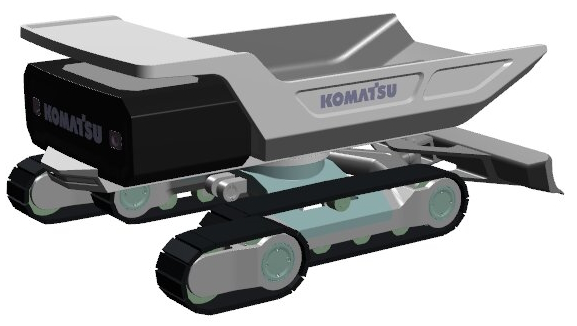}
    \caption{3D models of the excavator and the dump truck. Actuated joints are marked in green. 
    The machine width is 2.5 m.}
    \label{fig:machines}
\end{figure}

The machines are 3.3 m long and 2.5 m wide. The excavator has 3 ton mass,
and the empty dump truck weighs 2.0 tons.
The actuators can deliver torque up to a maximum value but the required torques
never reached these limits in the experiments.
The target drive speed of the machines was 0.3 m/s when carrying a load
and 0.35 m/s when driving empty.
The crawlers were modelled using the AGX Tracks module. The number
of track elements is about 50 and 20, for the main crawlers and sub-crawlers, respectively.
To support numerical stability at integration with 10 ms time-steps, strong
numerical viscous damping is enforced in the track element interconnection joints.
That results in excessive power consumption when driving the crawlers.
Therefore, the computed power was normalised by subtracting the dissipation 
observed when driving the tracks with no resistance from the terrain (having
the vehicle elevated above ground) and introducing an efficiency factor (constant). 

The lunar surface has an initial shape represented by an artificially generated
elevation map.  The terrain is given uniform soil properties, specifically the regolith 
was assigned a bank state internal friction angle of $0.80$ rad, cohesion $900$ Pa, 
dilatancy angle $0.23$ rad, mass density of $1580$ kg/m$^3$ at packing density $0.66$, 
and compression index of $0.11$. The local gravity acceleration is set to $1.6$ m/s$^2$.

\begin{figure}[H]
    \centering
    \includegraphics[width=0.65\textwidth]{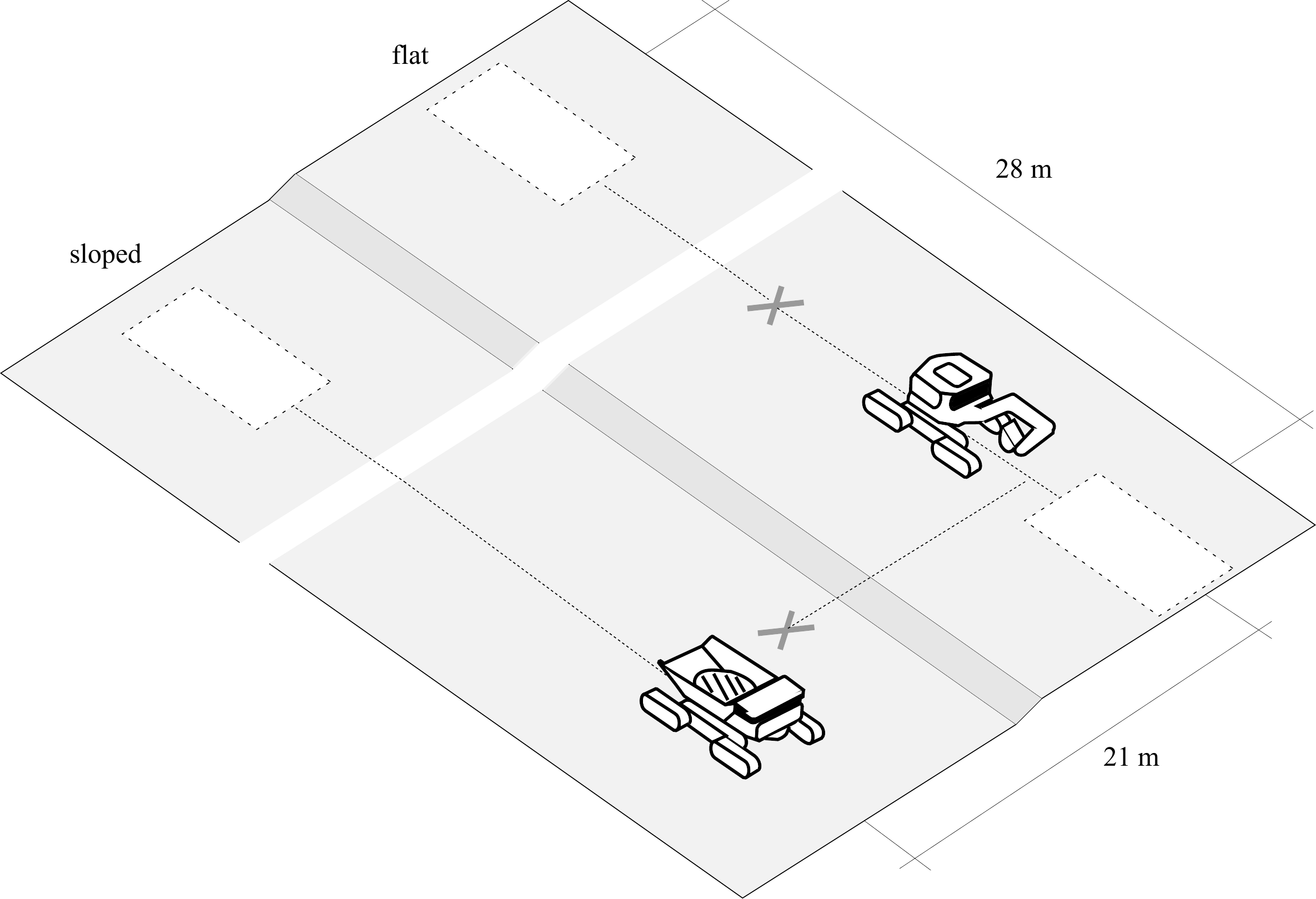}
    \caption{Illustration of Scenario 1 - Excavate lunar regolith from a dig area 
    (right dashed rectangle), offloaded on a dump truck that carries the material to
    a dump area (left dashed rectangle) where it is offloaded. Two dump areas are considered,
    separated by a slope.  The scales are not accurate to the real scene.}
    \label{fig:scenario-1}
\end{figure}

\subsection{Scenario 1 - Excavate, haul, and dump lunar regolith}
The construction instruction in this scenario was to repeatedly excavate and move soil
from a dig area to a dump area with boundaries specified by coordinates.
The excavator and dump truck have behaviour trees as described in the previous
section.



Two variations of the scenario were made, and we refer to them as \emph{flat} and
\emph{sloped}. They differ by the location of the dump area. In the flat case, the dump area is
located in the same region of rather flat terrain as the dig area. See Fig.~\ref{fig:scenario-1}
for an illustration. In the sloped
case, the dig and dump areas are separated by a slope that the excavator must cross
each digging cycle to reach the dump truck at the lower region.
The distance to reach the dump truck from the dig area is, on the other hand, 20\%
shorter on average in the sloped case.

The dump truck will start the dumping process when its target load is reached. When
driving to the dumping position, an extra waypoint will be created where the dump truck will
get the correct dumping orientation. It will then drive straight to the final position. This is to
avoid unwanted interaction between the pile and the dump truck. When dumping, the
dump bed will rest in a raised position for a couple of seconds. The dump truck will then
move forward with the bed still raised before lowering the bed and driving towards a
loading position.

\begin{figure}[H]
    \centering
    \includegraphics[width=0.45\textwidth,trim={140mm 75mm 140mm 80mm},clip]{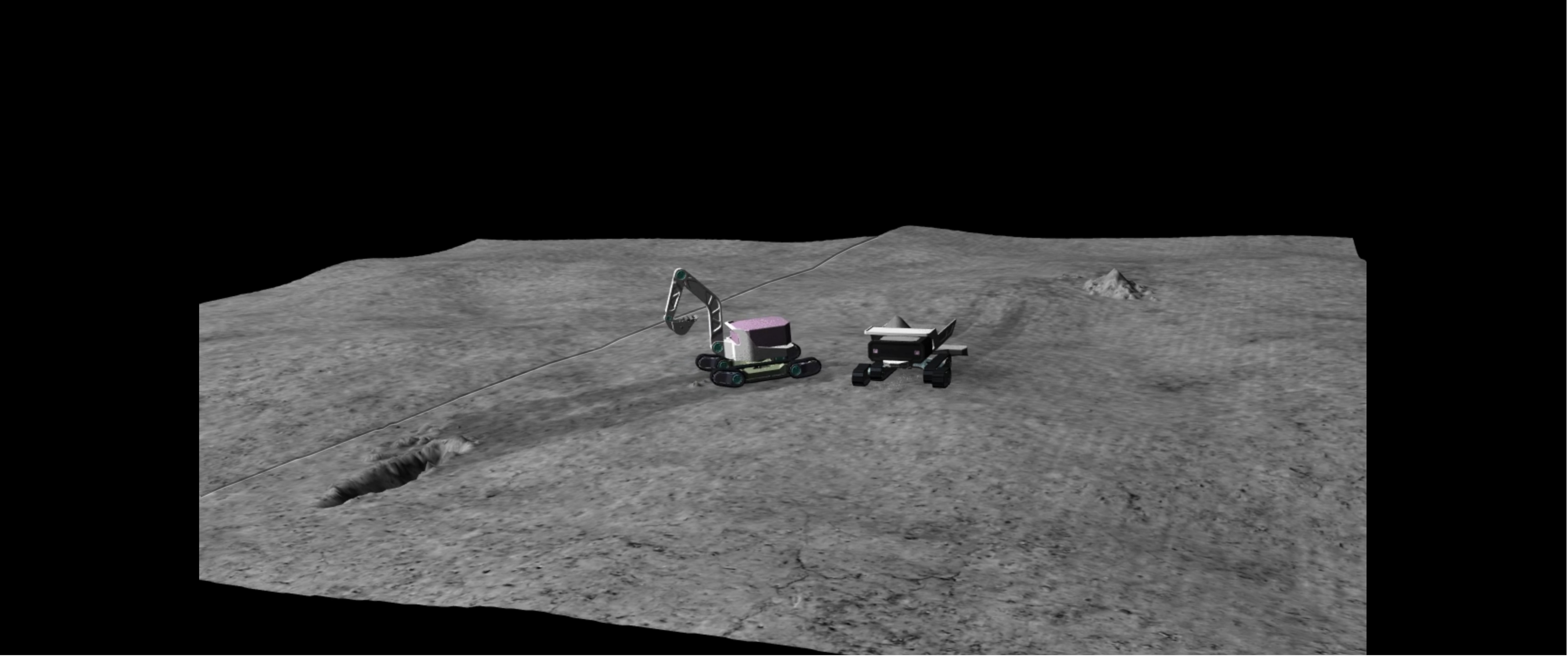}
    \includegraphics[width=0.45\textwidth,trim={120mm 75mm 160mm 80mm},clip]{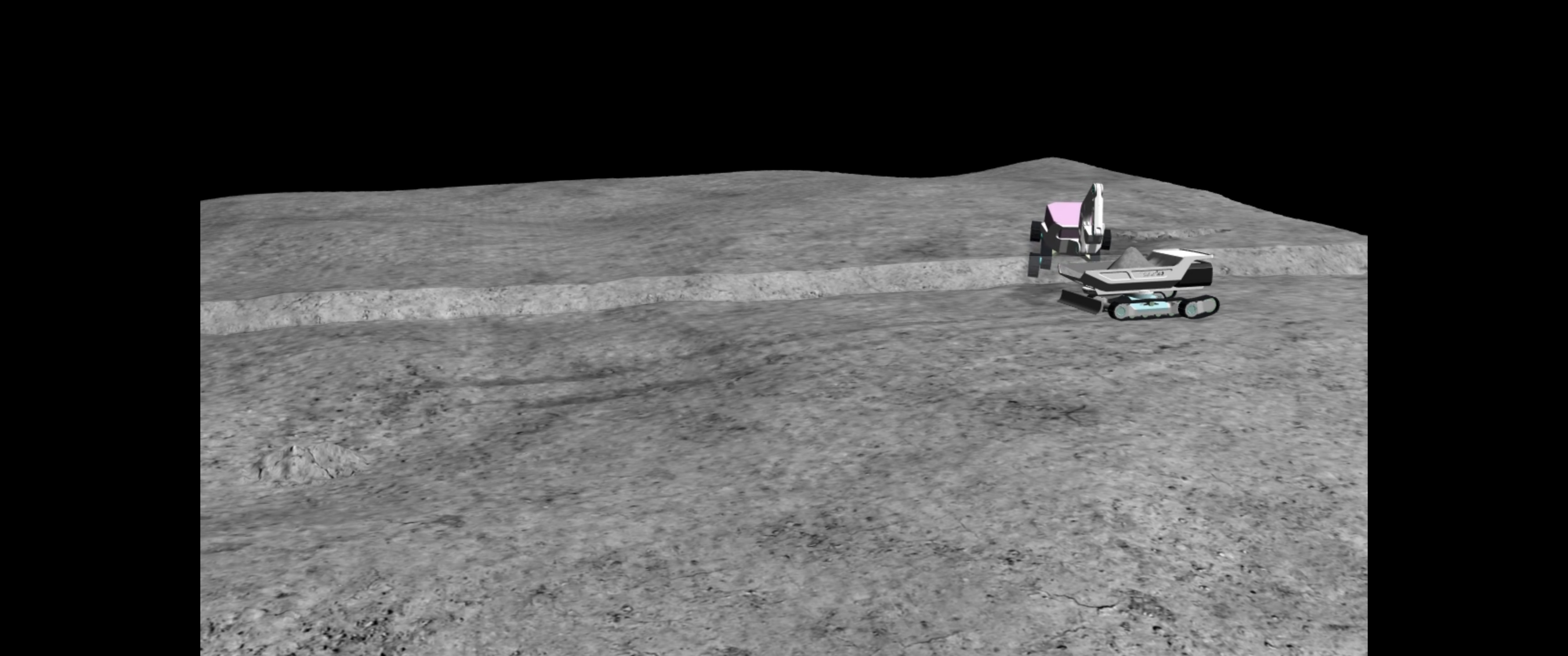}\\
    \includegraphics[width=0.45\textwidth,trim={140mm 75mm 140mm 80mm},clip]{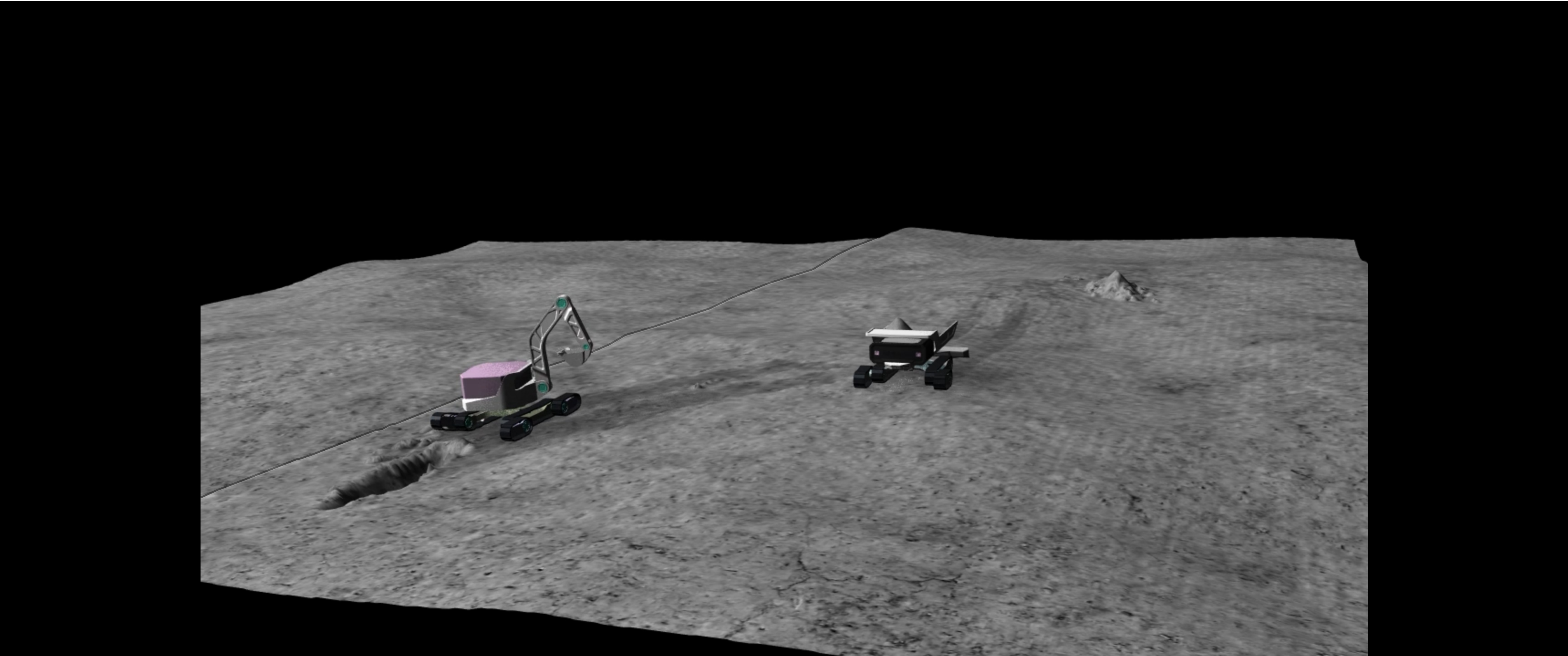}
    \includegraphics[width=0.45\textwidth,trim={120mm 75mm 160mm 80mm},clip]{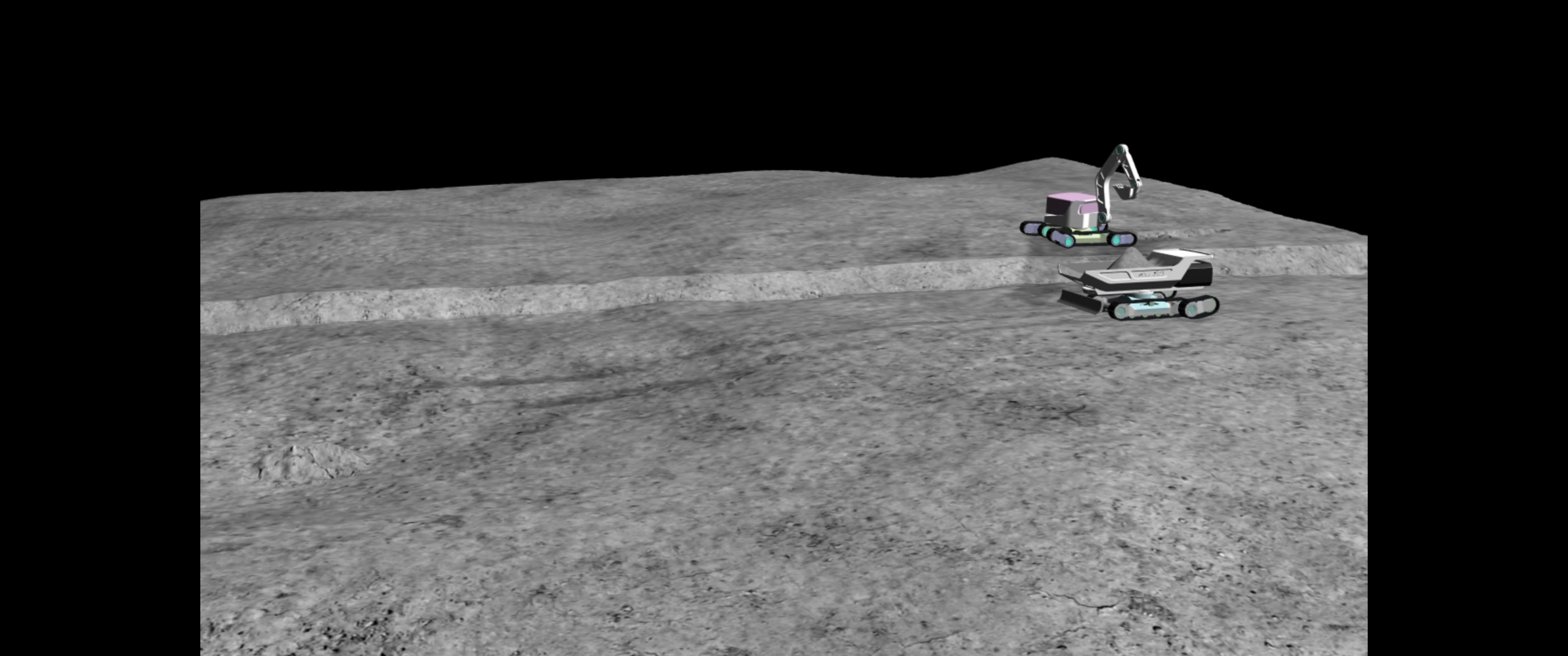}\\
    \includegraphics[width=0.45\textwidth,trim={140mm 75mm 140mm 80mm},clip]{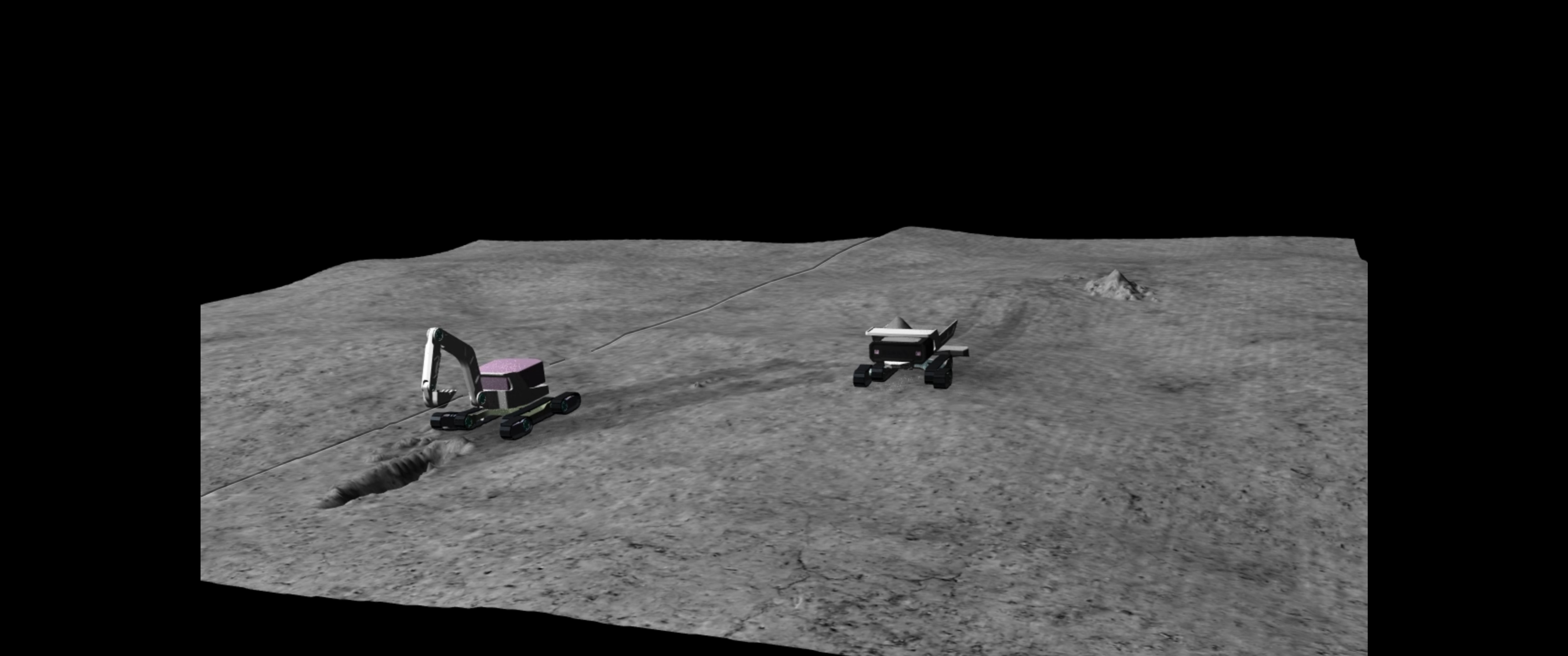}
    \includegraphics[width=0.45\textwidth,trim={120mm 75mm 160mm 80mm},clip]{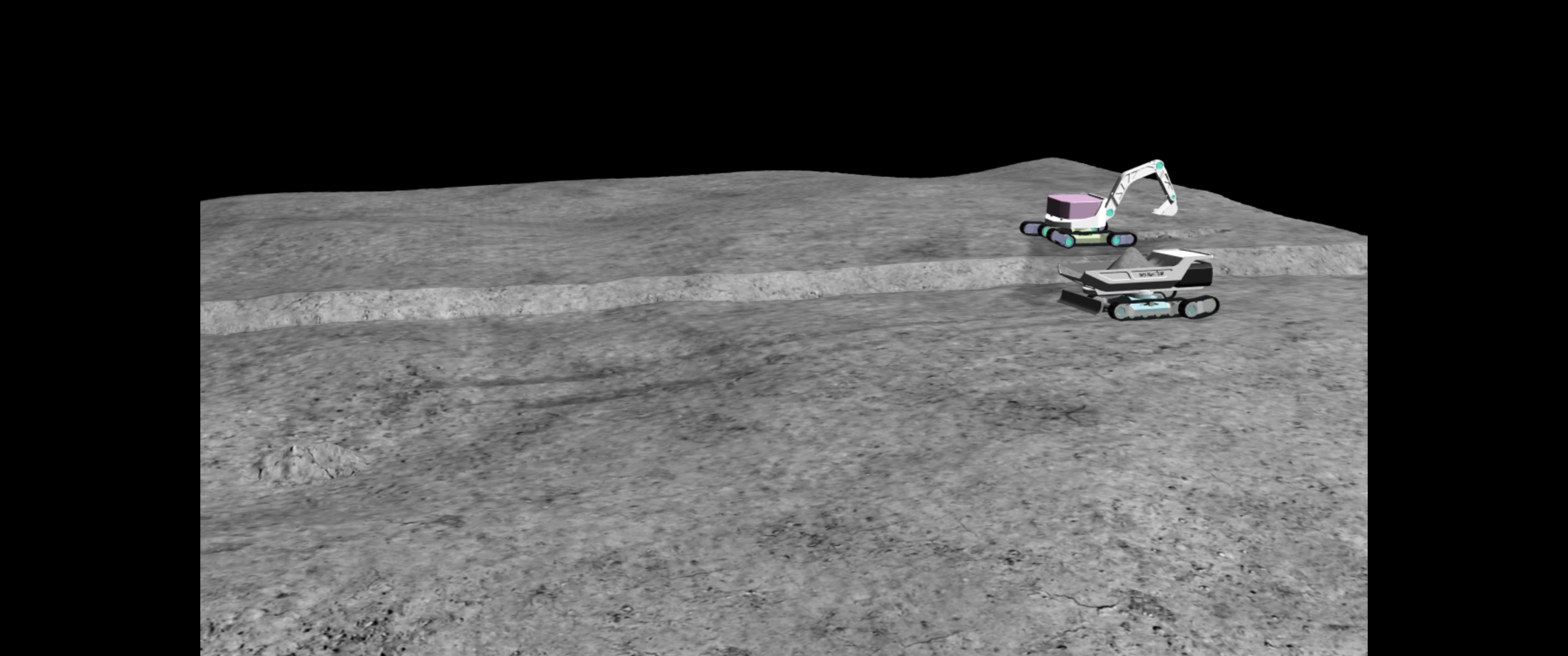}\\
    \includegraphics[width=0.45\textwidth,trim={140mm 75mm 140mm 80mm},clip]{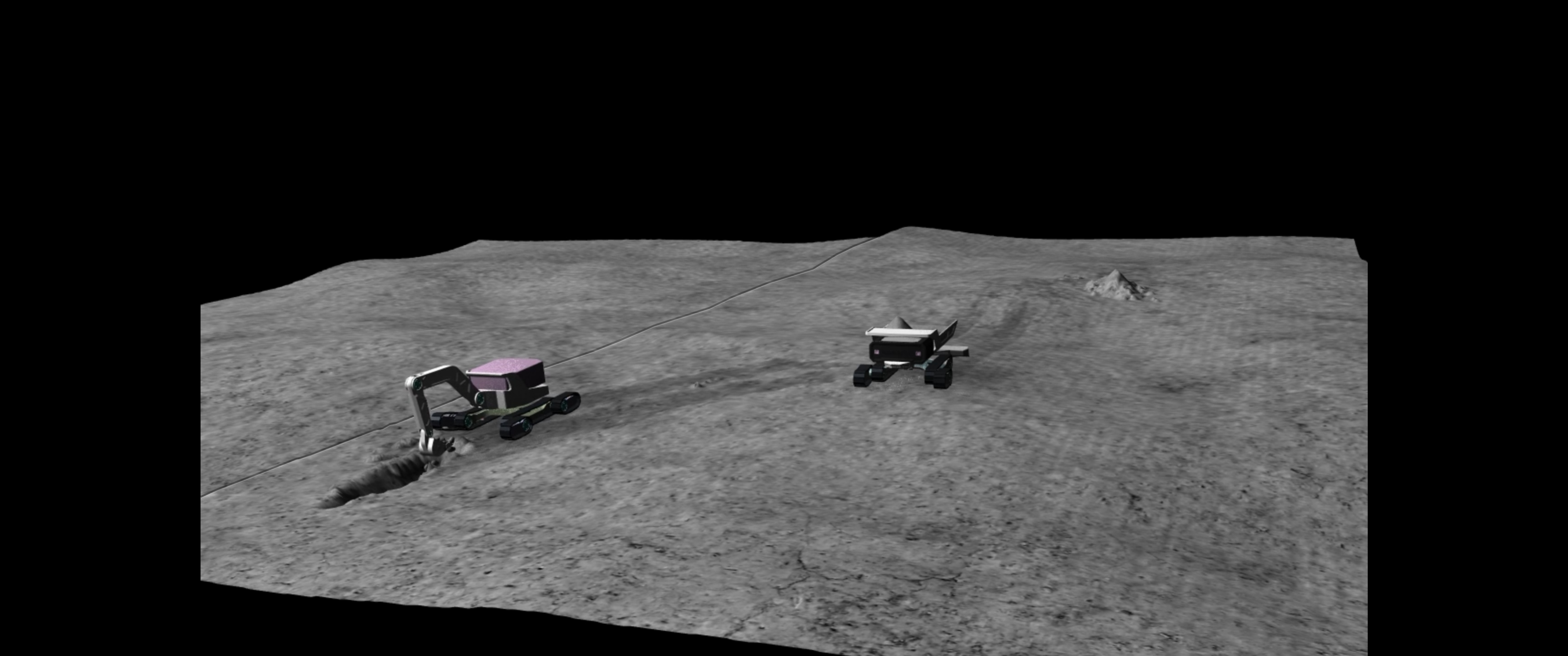}
    \includegraphics[width=0.45\textwidth,trim={120mm 75mm 160mm 80mm},clip]{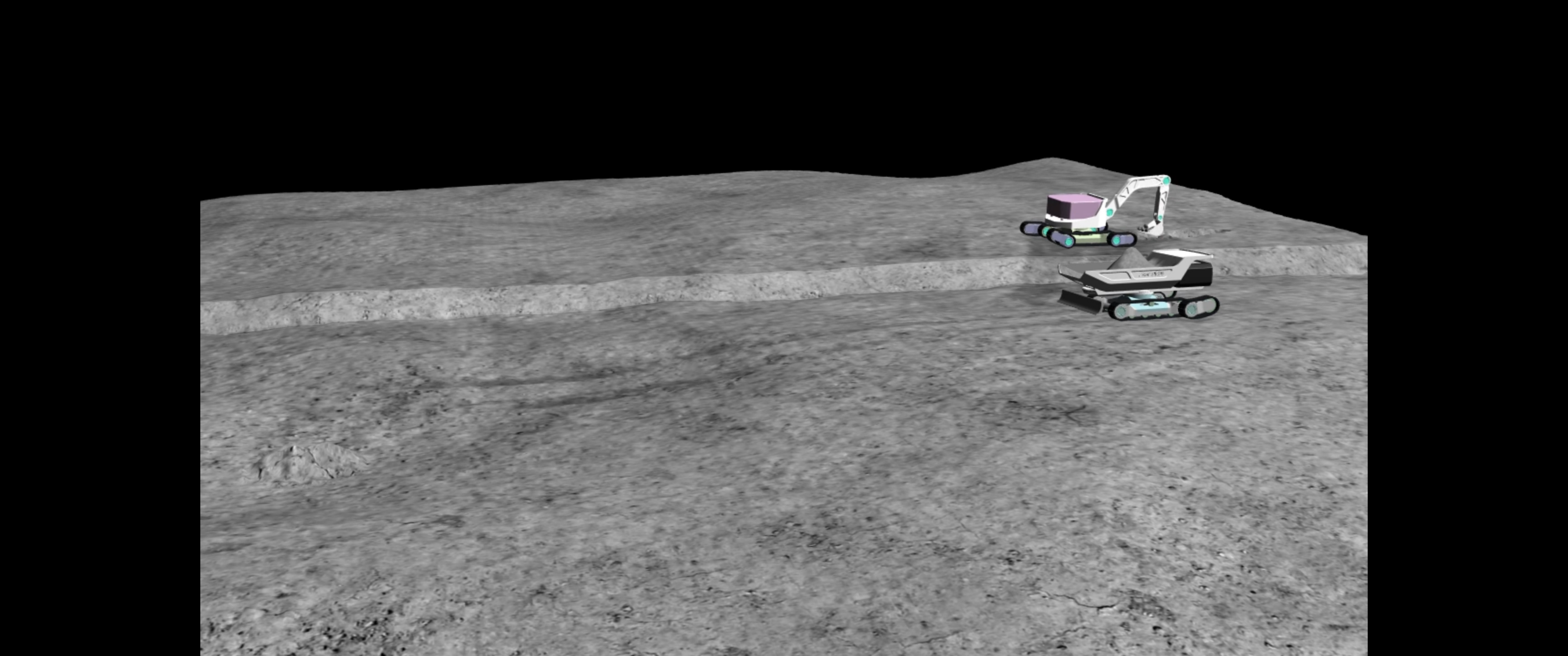}\\
    \includegraphics[width=0.45\textwidth,trim={140mm 75mm 140mm 80mm},clip]{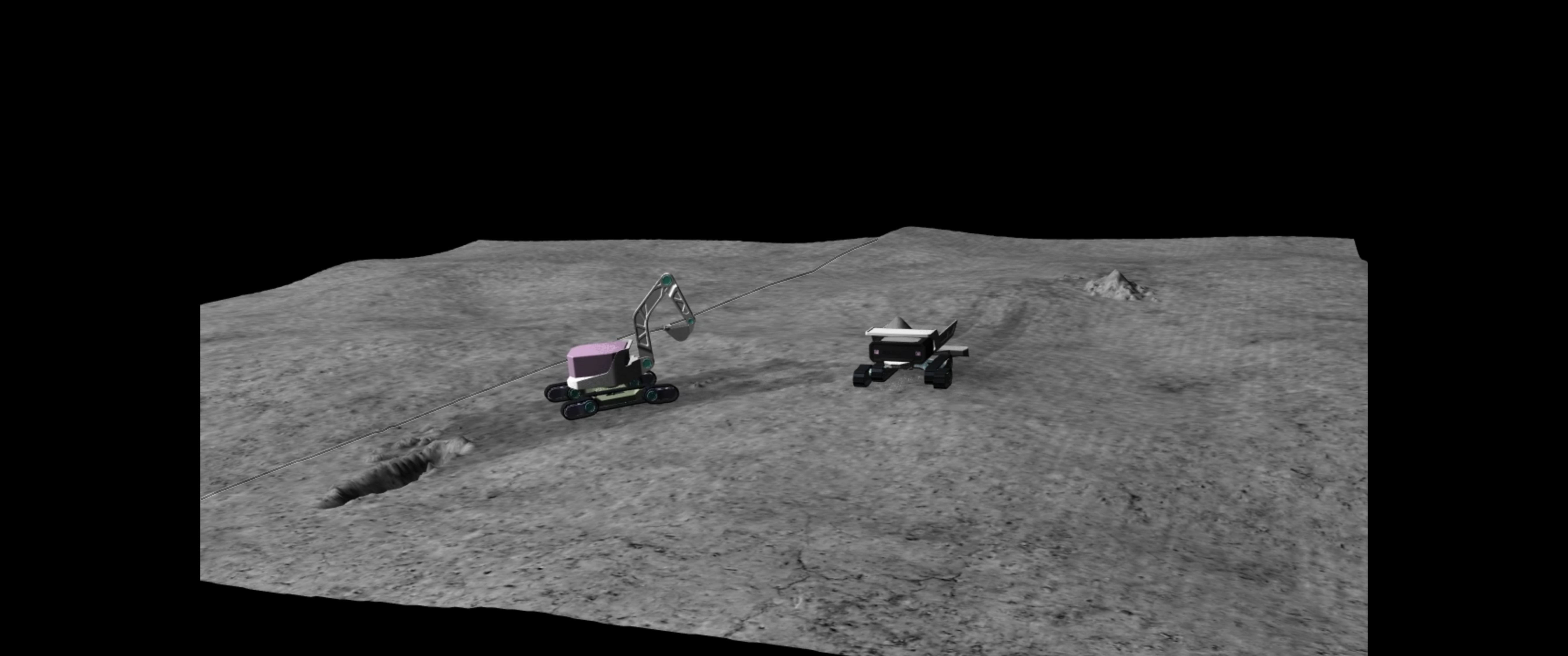}
    \includegraphics[width=0.45\textwidth,trim={120mm 75mm 160mm 80mm},clip]{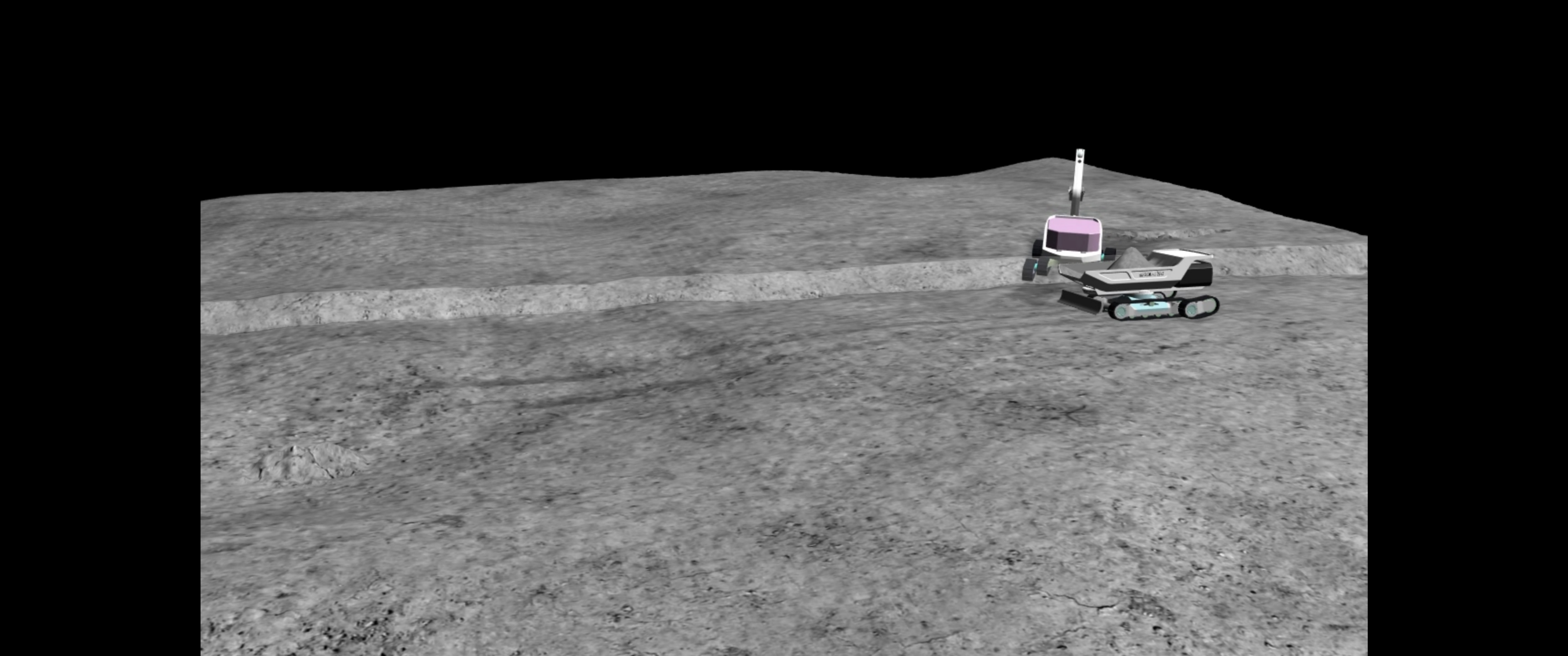}\\
    \includegraphics[width=0.45\textwidth,trim={140mm 75mm 140mm 80mm},clip]{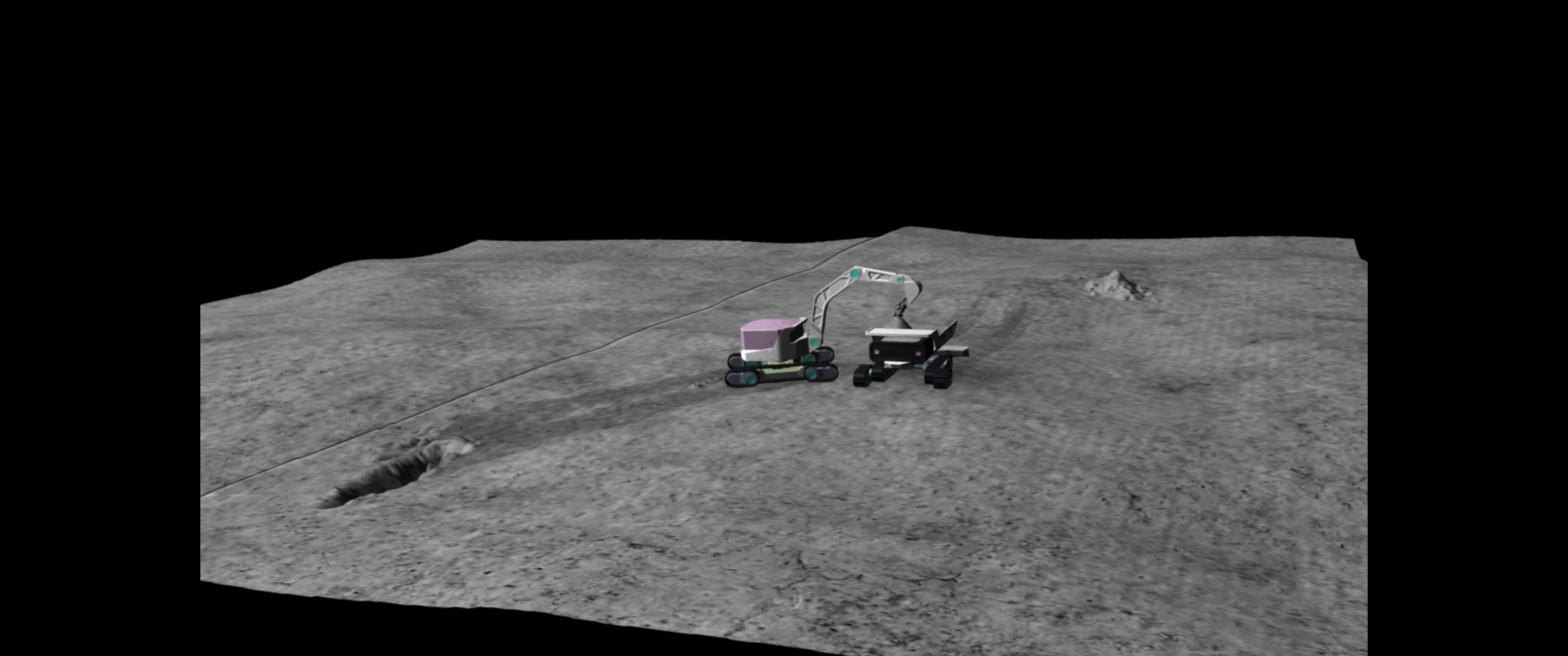}
    \includegraphics[width=0.45\textwidth,trim={120mm 75mm 160mm 80mm},clip]{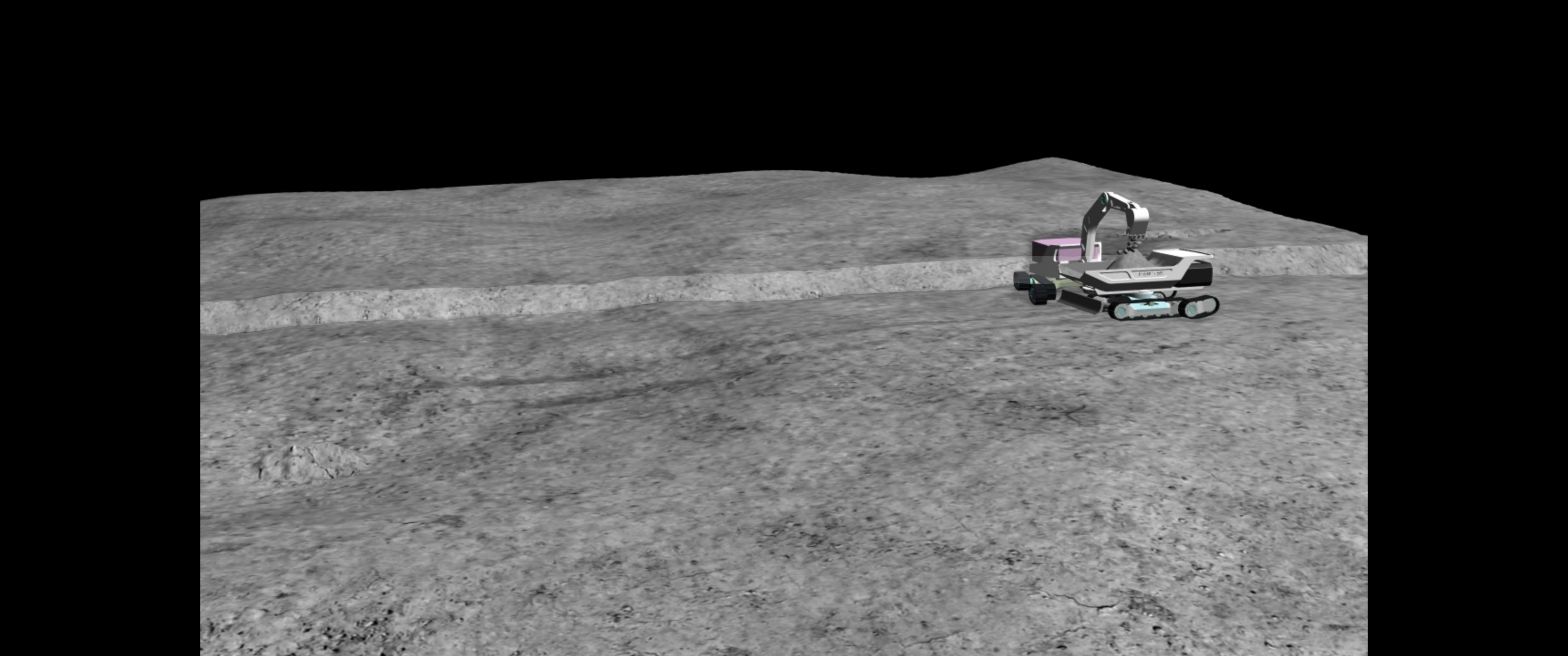}\\
    \includegraphics[width=0.45\textwidth,trim={120mm 75mm 160mm 80mm},clip]{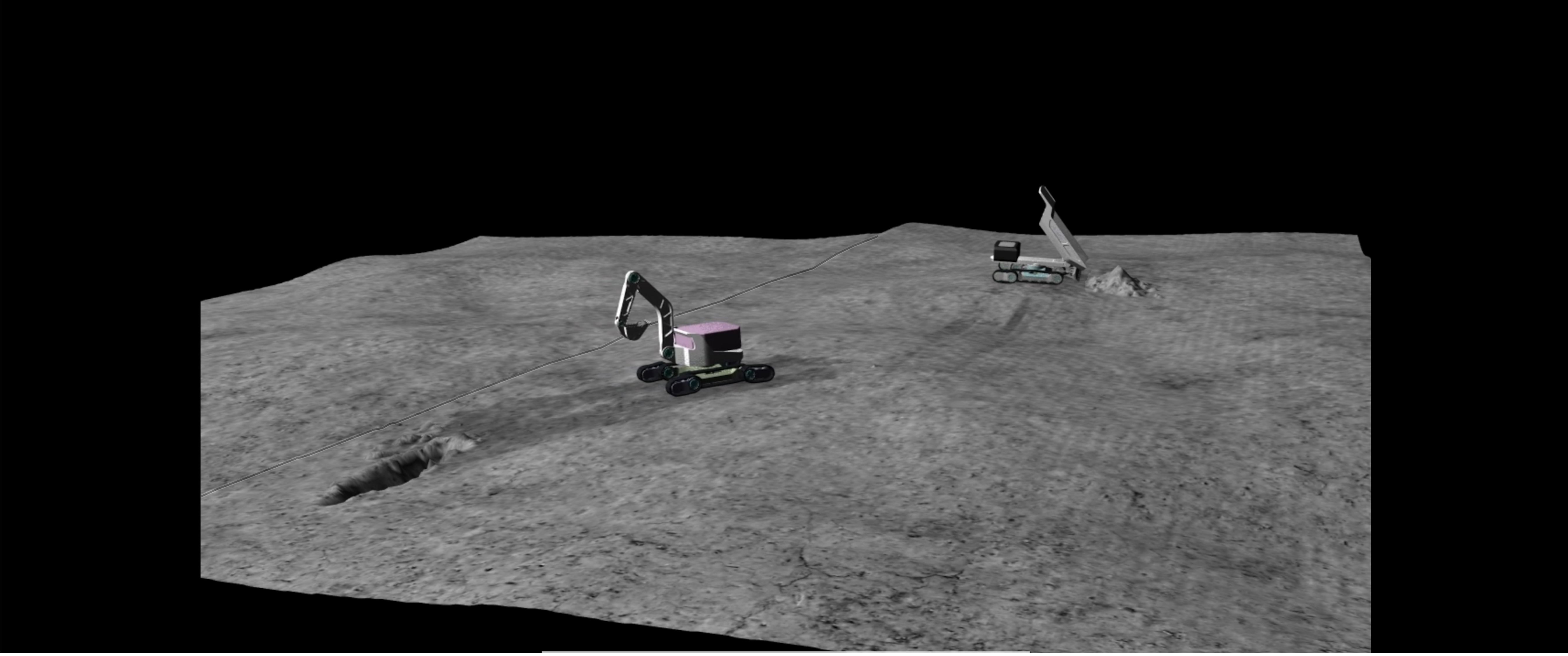}
    \includegraphics[width=0.45\textwidth,trim={120mm 75mm 160mm 80mm},clip]{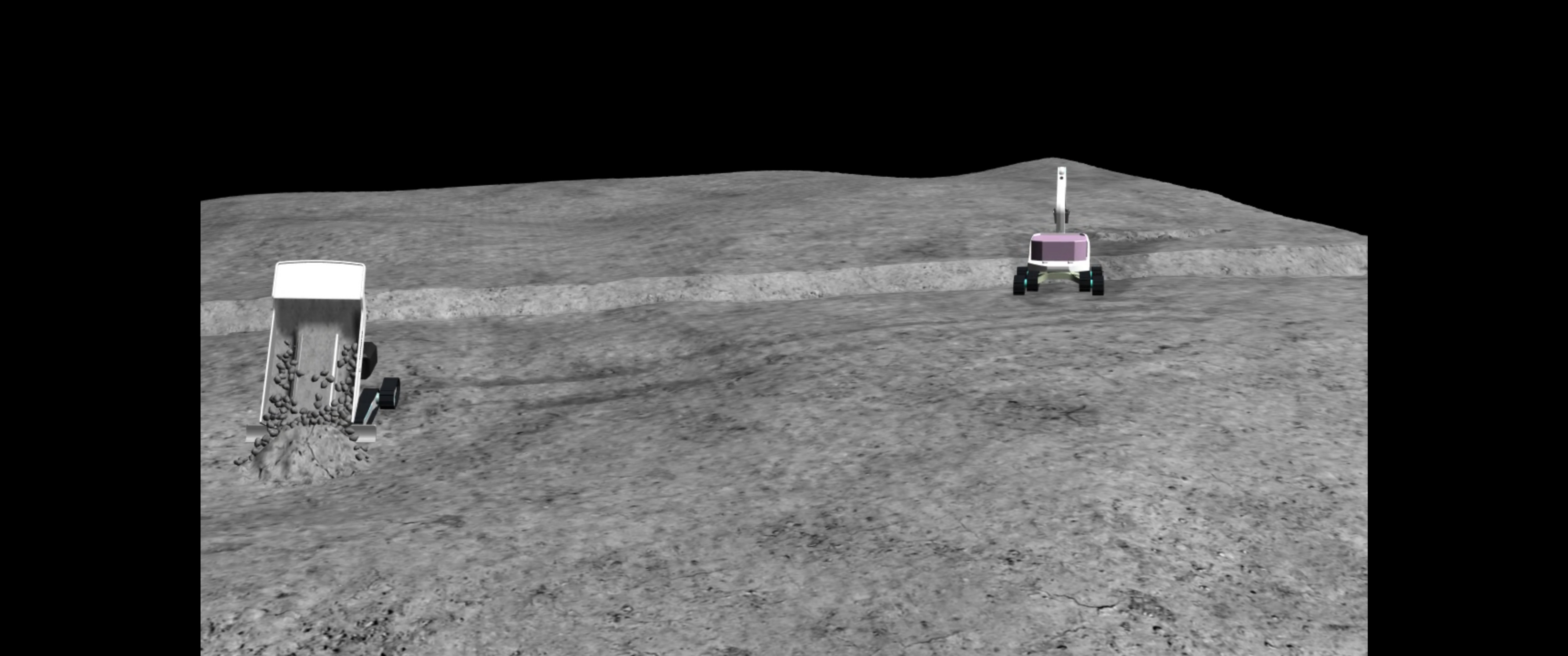}
    \caption{Image sequence from one excavation cycle in the scenario \texttt{Excavation-flat} (left) 
    and \texttt{Excavation-slope} (right). The sequence ends with the crawler dumping its load at a dump area.}
    \label{fig:sequence_flat_sloped}
\end{figure}

The construction work was simulated over 30 digging cycles, and virtual sensor data
was collected, including time series for vehicle pose, kinematics, actuator torques, 
and loaded and dumped soil mass. Power consumption was computed by integrating the (positive) work exerted by the actuators.
Image sequences from the two simulations can be found in Fig.~\ref{fig:sequence_flat_sloped}.

A summary of the results is found in Table.~\ref{table:scenario1}. The average loaded
mass was 8\% higher in the flat case. This is attributed to the fact that it is more 
difficult to track the excavator’s planned path across the slope than over the flat terrain. 
Consequently, the excavator sometimes ends up in a skewed pose where it is harder to fill the bucket.
In both cases, there is material spilling (6-7\%) from the bucket.
This occurs mainly when turning the chassis, apparently causing too large peaks in acceleration.
The average cycle times are very similar but the power consumption is 14\% higher for the 
sloped case due to the slope traversal.

\begin{table}
    \caption{Mean and standard deviation for loaded and spilt mass, duration, and work
    over 30 cycles in Scenario 1 on flat and sloped terrain.}
    \label{table:scenario1}
    {\small
    \begin{tabular}{|l|c|c|c|c|}\hline
        Case    &   Mass (kg)  &    Spill (kg)  &  Time (s)     &   Work (kJ)   \\\hline
        Flat        &   $69\pm26$  &    $5\pm7$     &  $117\pm13$   &   $66\pm16$   \\\hline
        Sloped      &   $64\pm24$  &    $4\pm5$     &  $118\pm15$   &   $74\pm24$   \\\hline
    \end{tabular}
    }
\end{table}

\begin{figure}[H]
    \centering
    \subfloat[flat]{
        \centering
        \includegraphics[width=0.48\textwidth,trim={0mm 0mm 0mm 0mm},clip]{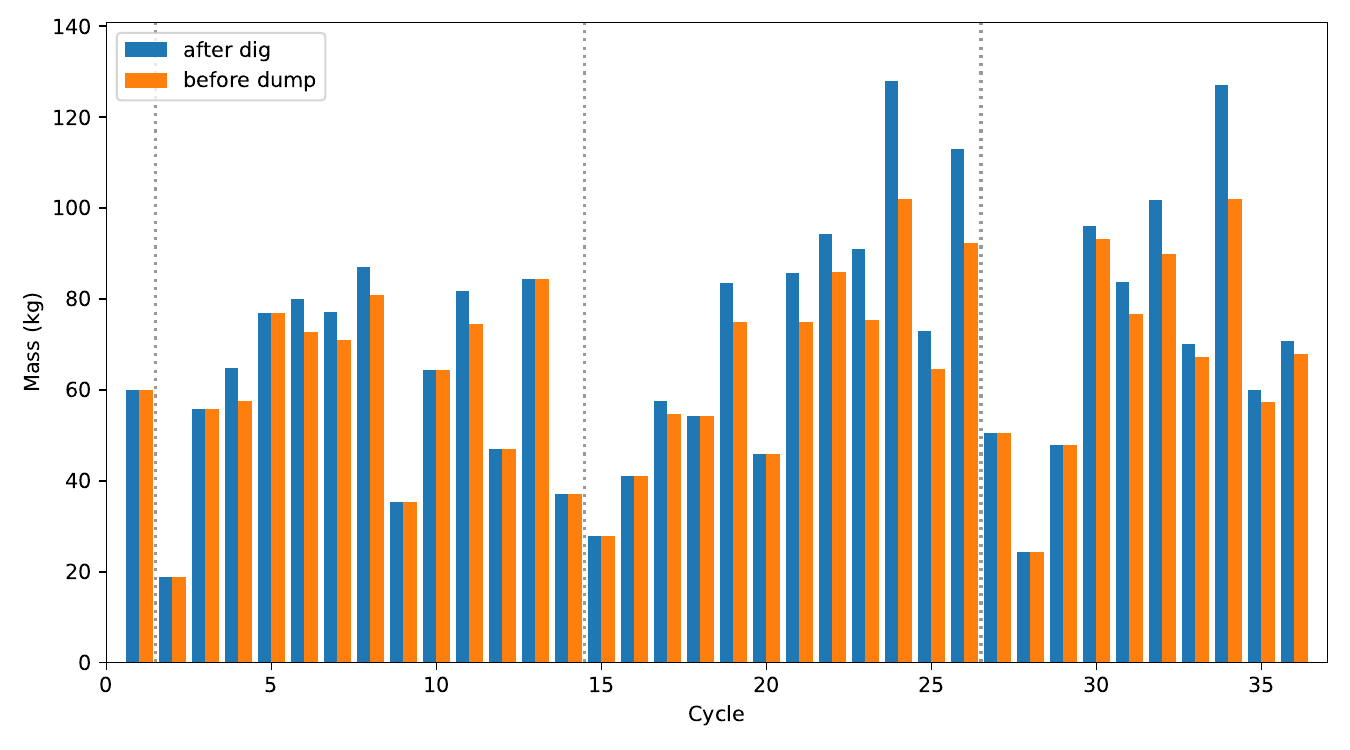}
        }
    \subfloat[sloped]{
        \centering
        \includegraphics[width=0.48\textwidth,trim={0mm 0mm 0mm 0mm},clip]{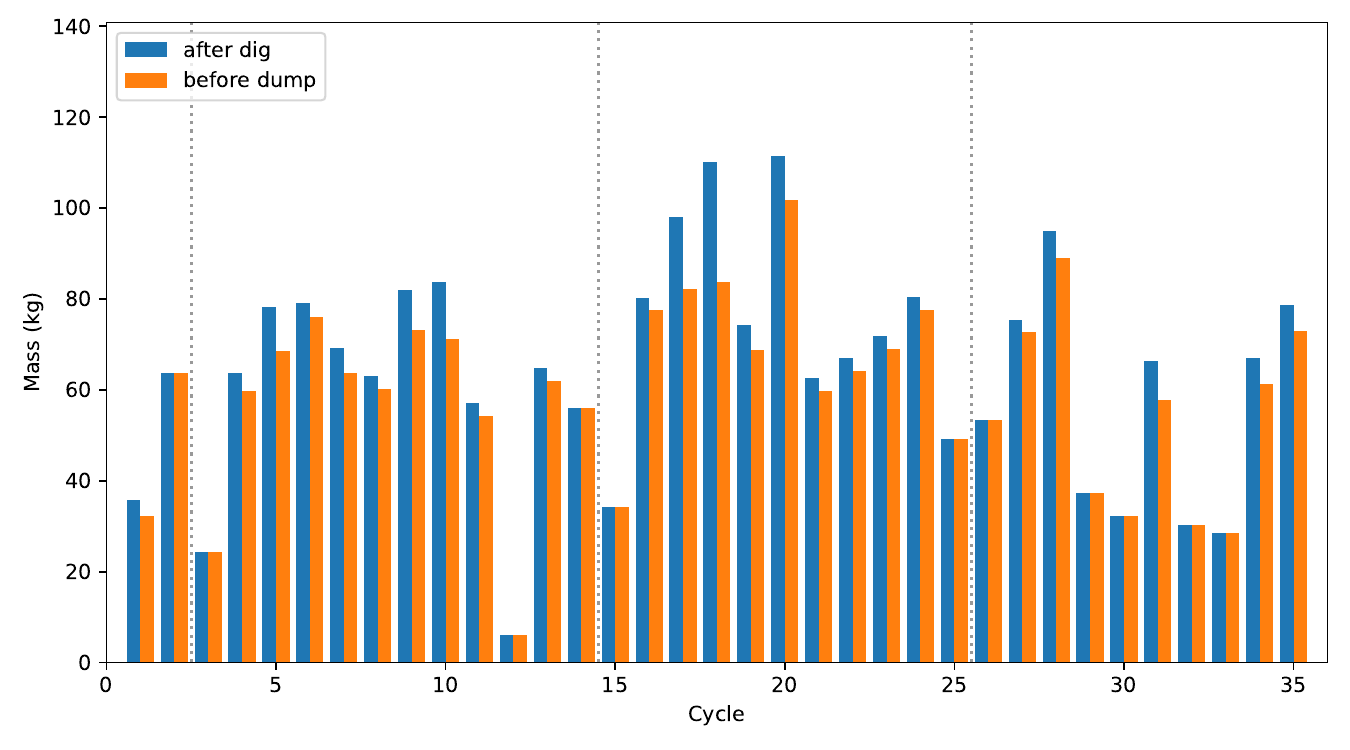}
        }
    \caption{Evolution of the excavated and dumped mass over 30 cycles for the two cases of Scenario 1.}
    \label{fig:mass}
\end{figure}

\begin{figure}[H]
    \centering
        \subfloat[flat]{
        \centering
        \includegraphics[width=0.48\textwidth,trim={0mm 0mm 0mm 0mm},clip]{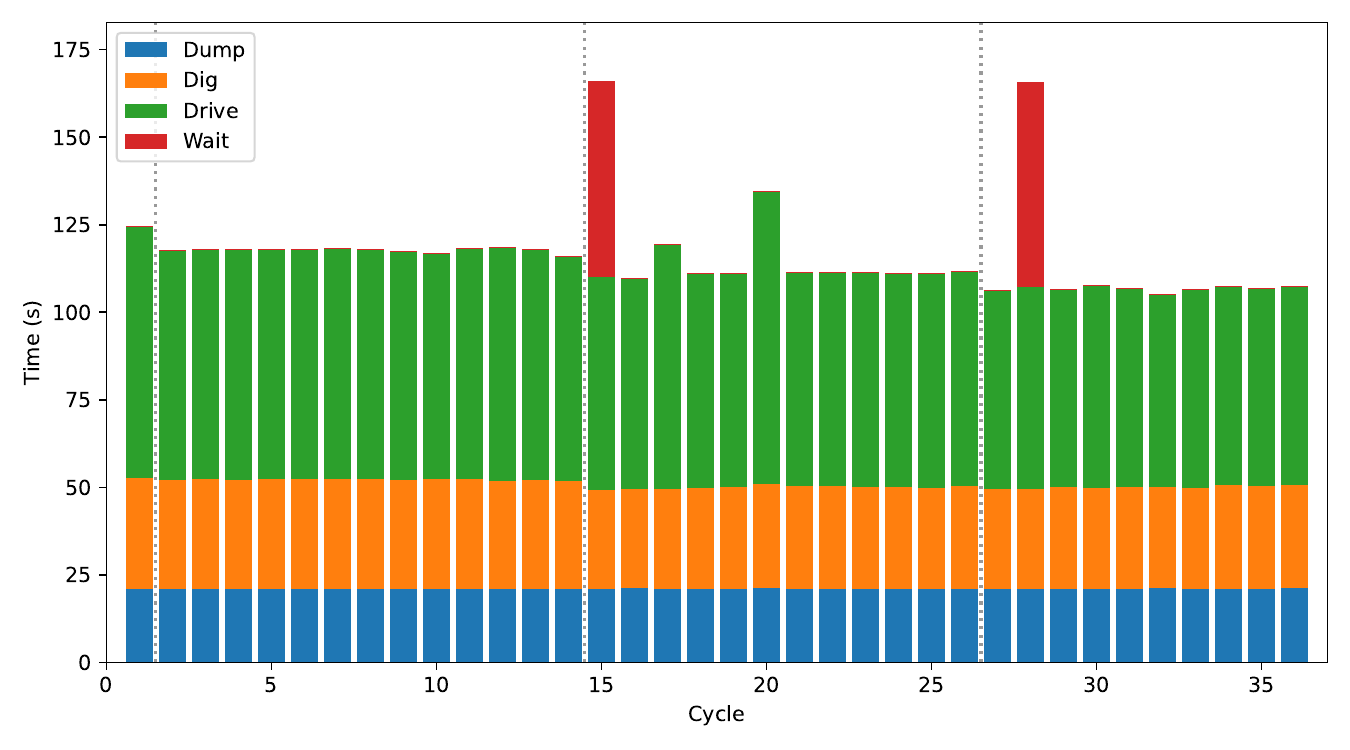}
        }
    \subfloat[sloped]{
        \centering
        \includegraphics[width=0.48\textwidth,trim={0mm 0mm 0mm 0mm},clip]{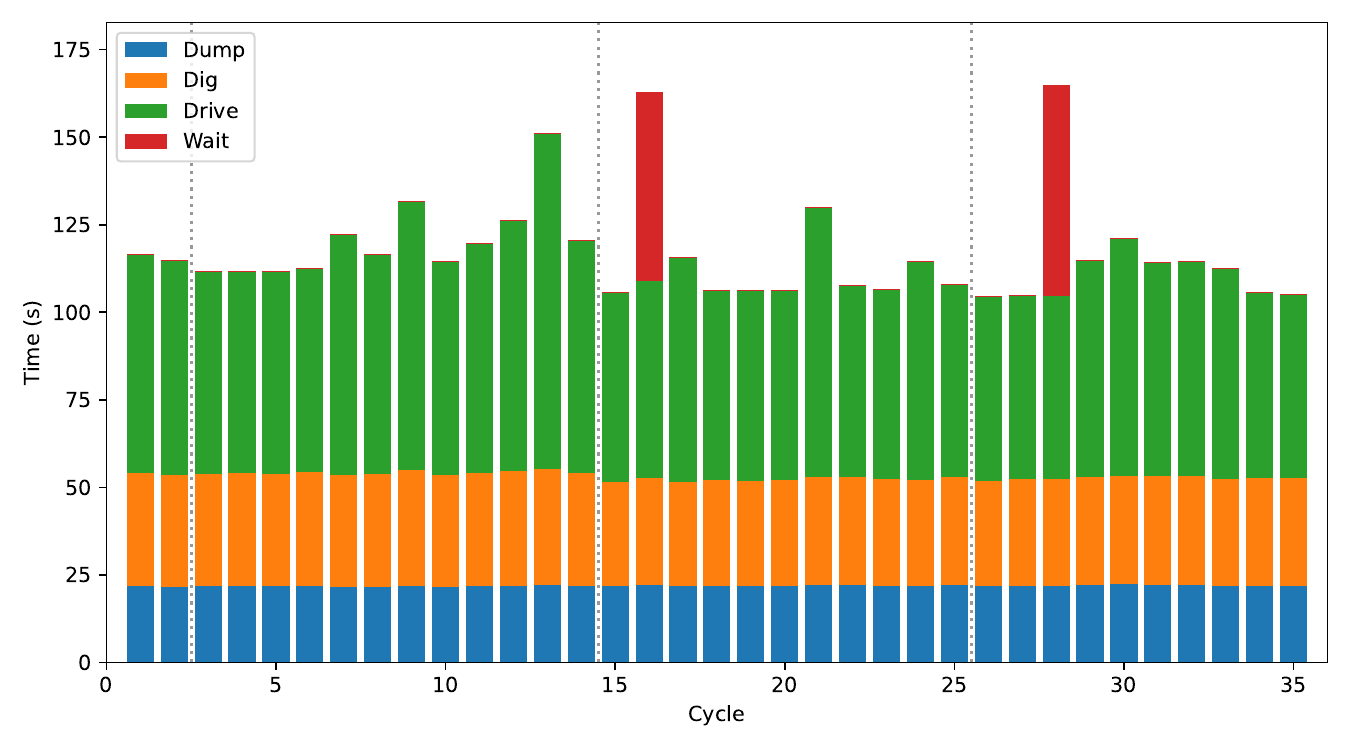}
        }
    \caption{Evolution of the working time over 30 cycles for the two cases of Scenario 1.}
    \label{fig:duration}
\end{figure}

\begin{figure}[H]
    \centering
        \subfloat[flat]{
        \centering
        \includegraphics[width=0.48\textwidth,trim={0mm 0mm 0mm 0mm},clip]{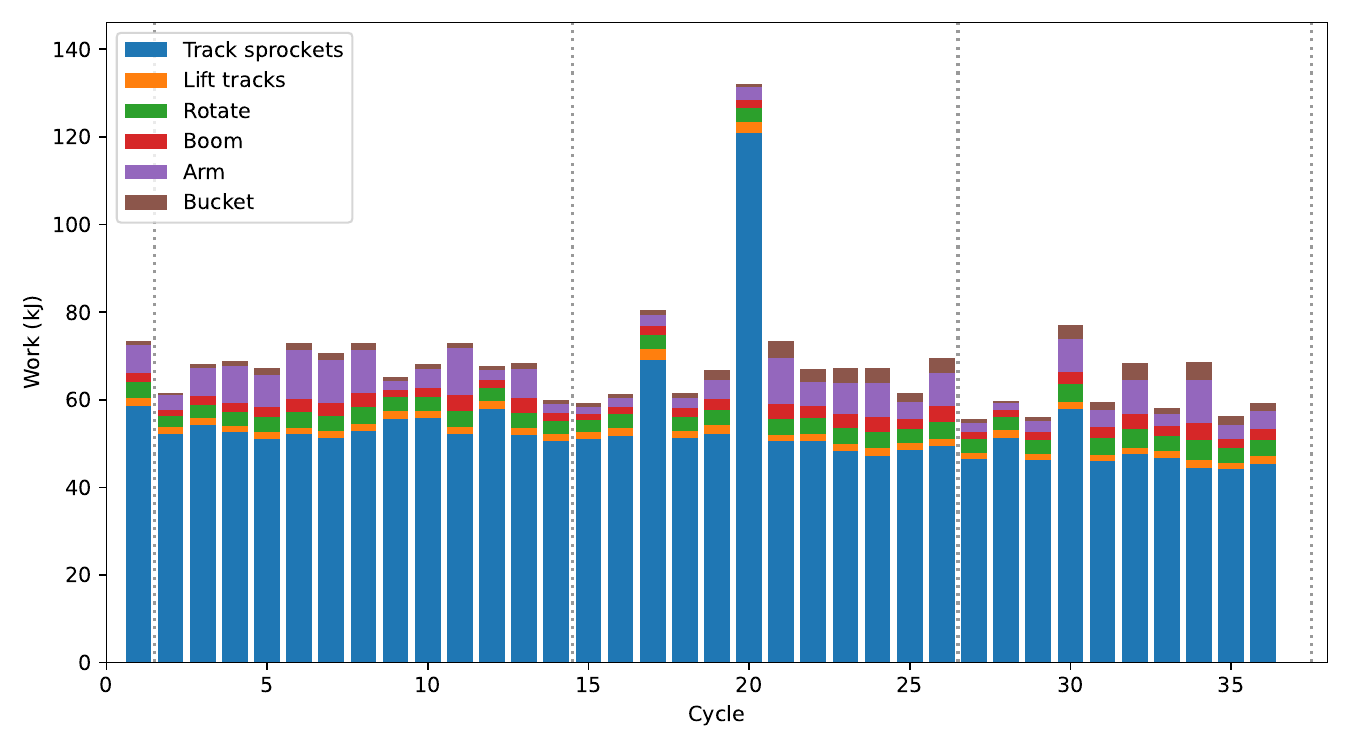}
        }
        \subfloat[flat]{
        \centering
        \includegraphics[width=0.48\textwidth,trim={0mm 0mm 0mm 0mm},clip]{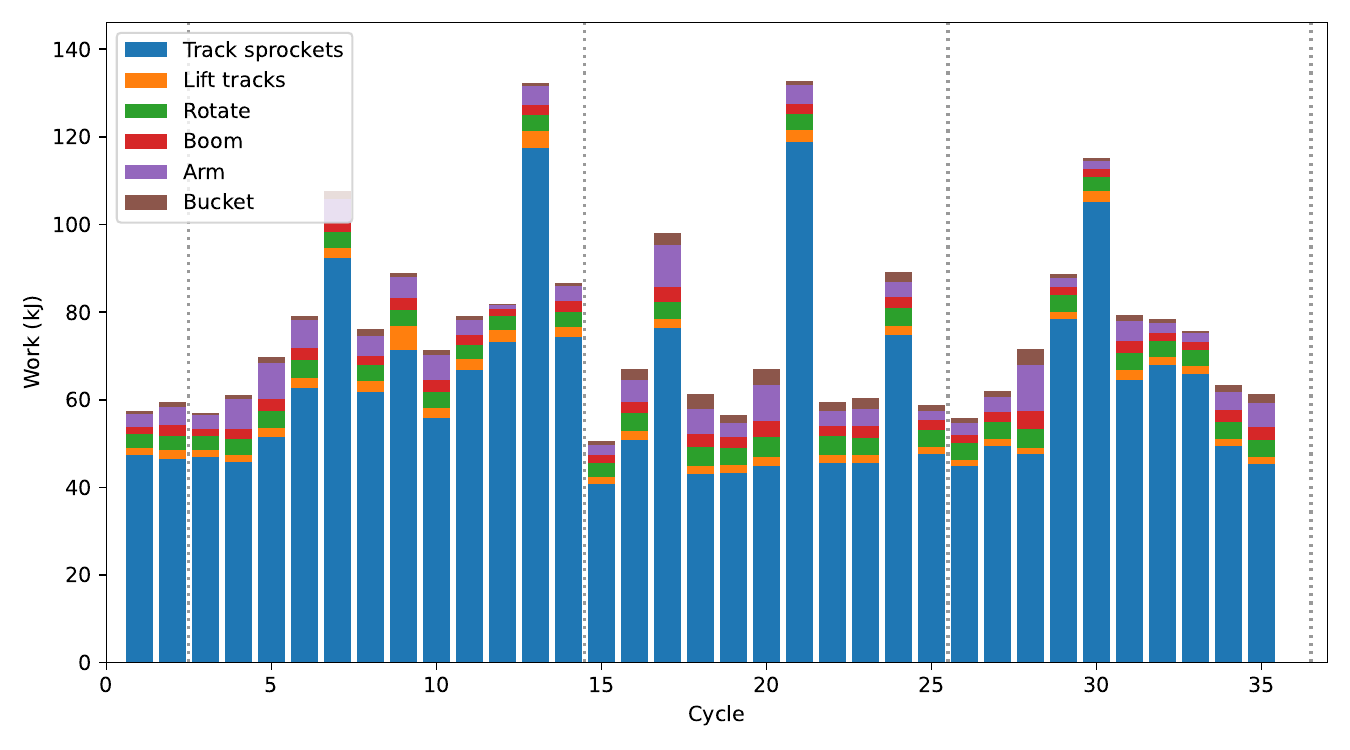}
        }
    \caption{Evolution of the work over 30 cycles for the two cases of Scenario 1.}
    \label{fig:work}
\end{figure}

\begin{figure}[H]
    \centering
        \subfloat[flat]{
        \centering
        \includegraphics[width=0.48\textwidth,trim={0mm 0mm 0mm 0mm},clip]{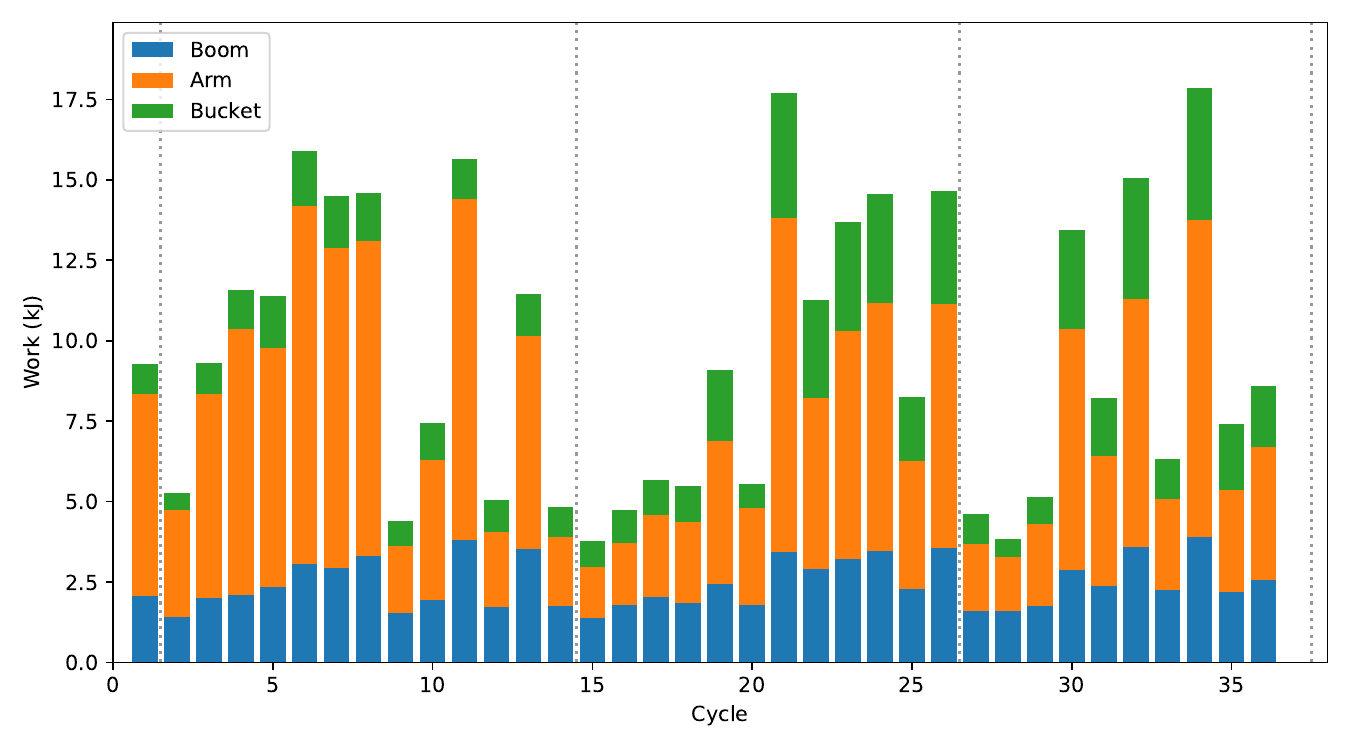}
        }
        \subfloat[flat]{
        \centering
        \includegraphics[width=0.48\textwidth,trim={0mm 0mm 0mm 0mm},clip]{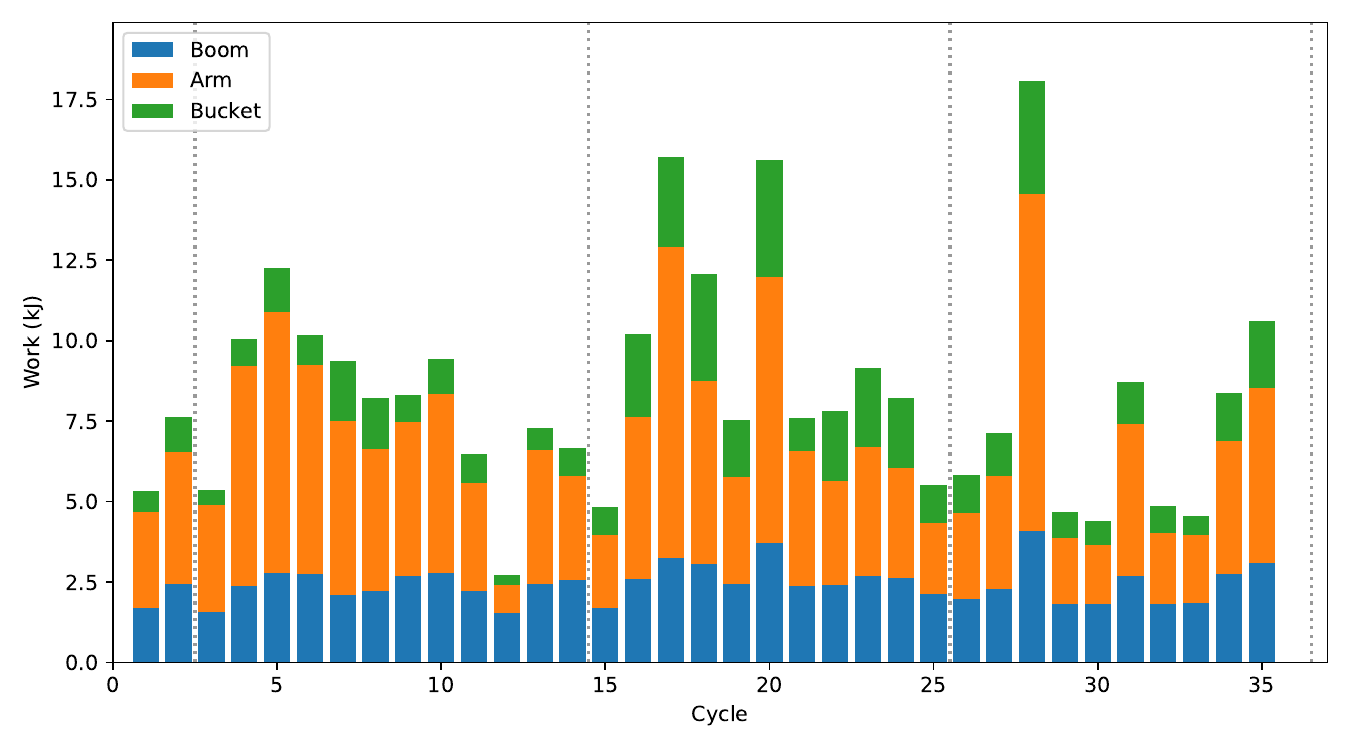}
        }
    \caption{Evolution of the work for the excavation process only over 30 cycles for the two cases of Scenario 1.}
    \label{fig:work-excavation}
\end{figure}

The evolution of the excavated and dumped mass, working time, and work per cycle are
displayed in Fig.~\ref{fig:mass}, \ref{fig:duration}, and \ref{fig:work},
respectively.  The moments when the excavator switches to a new cell are
marked with dashed lines. We observe that the mass (and power consumption) drops
at the switching events. This indicates that there is room for improving the 
planning to maximize the equipment utilization.
The working time for different tasks is nearly constant, with the exception 
of when the excavator waits for the dump truck to return.
The power consumption is dominated by running the crawlers although it was 
rescaled for the excessive dissipation mechanism for numerical stabilization.
As can be more clearly seen in Fig.~\ref{fig:work-excavation}, the arm actuator 
dominates the power consumption during digging and varies a lot
over the cycles, which can be understood by the fact that the planned bucket trajectory adapts
to the shape of the terrain, which evolves over time.
We hypothesize that the greater power consumption in the sloped case is due to the fact that
more uneven terrain leads to more situations where it is difficult, or even impossible,
not to slip significantly with the large tracks.

\subsection{Scenario 2 - Creating a predefined ground construction for habitat modules}

\begin{figure}[H]
    \centering
    \includegraphics[width=0.65\textwidth,trim={0mm 0mm 0mm 0mm},clip]{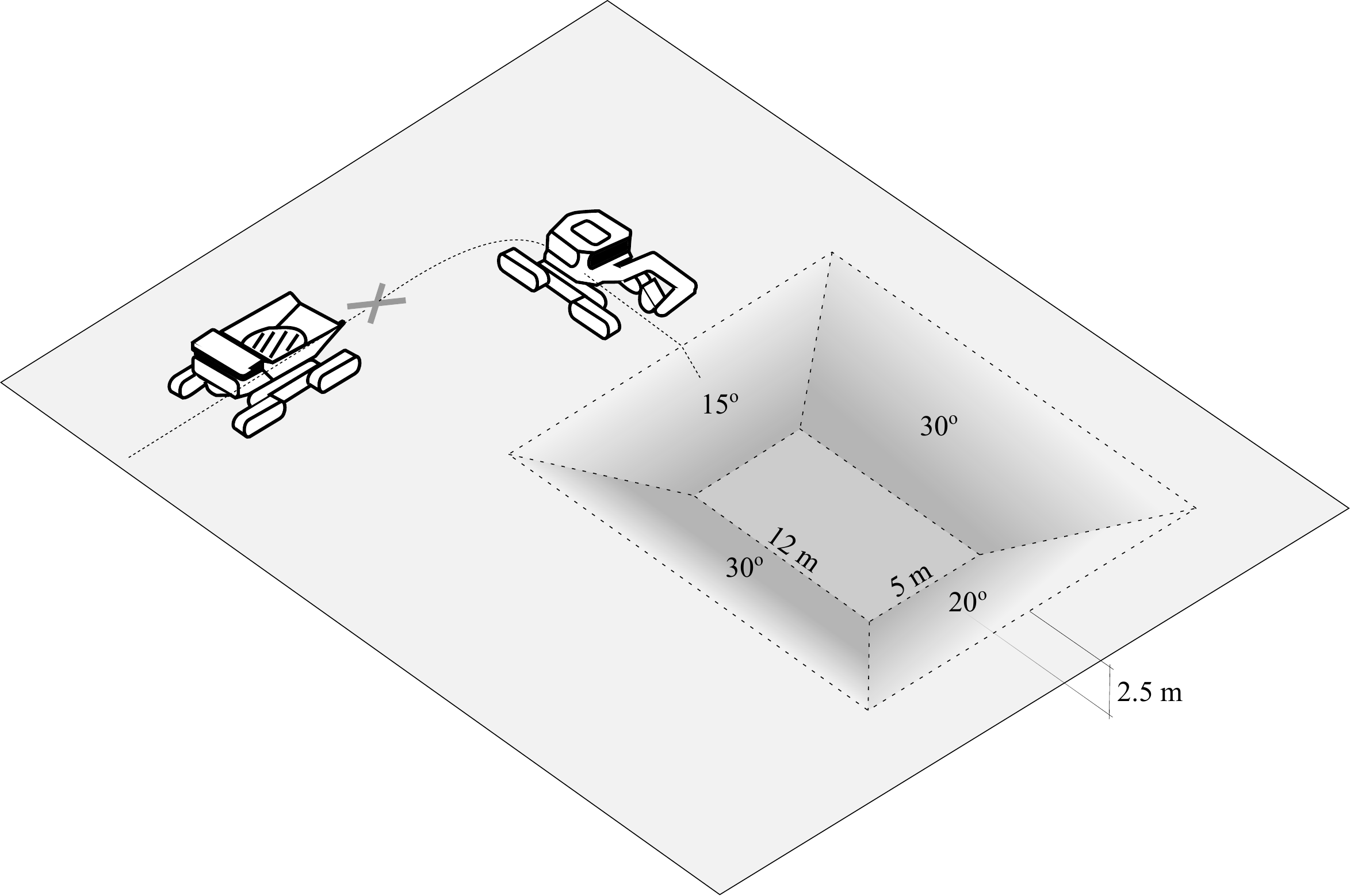}
    \caption{Illustration of Scenario 2 - Creating a predefined ground construction for 
    habitat modules by repeated excavation and final grading using the dump truck blade.}
    \label{fig:scenario-2}
\end{figure}

In this scenario, an excavator and a dump truck collaborate to shape the terrain
towards a specified 3D profile where habitat modules are to be placed and 
partially covered with regolith. The dump truck must adapt the location for parking to 
both the current dig location and the already excavated area in order to be easily 
accessible to the cycling excavator. See Fig.~\ref{fig:scenario-2} for an illustration.

\begin{figure}[H]
    \centering
    \includegraphics[height=0.2\textwidth,trim={200mm 0mm 0mm 120mm},clip]{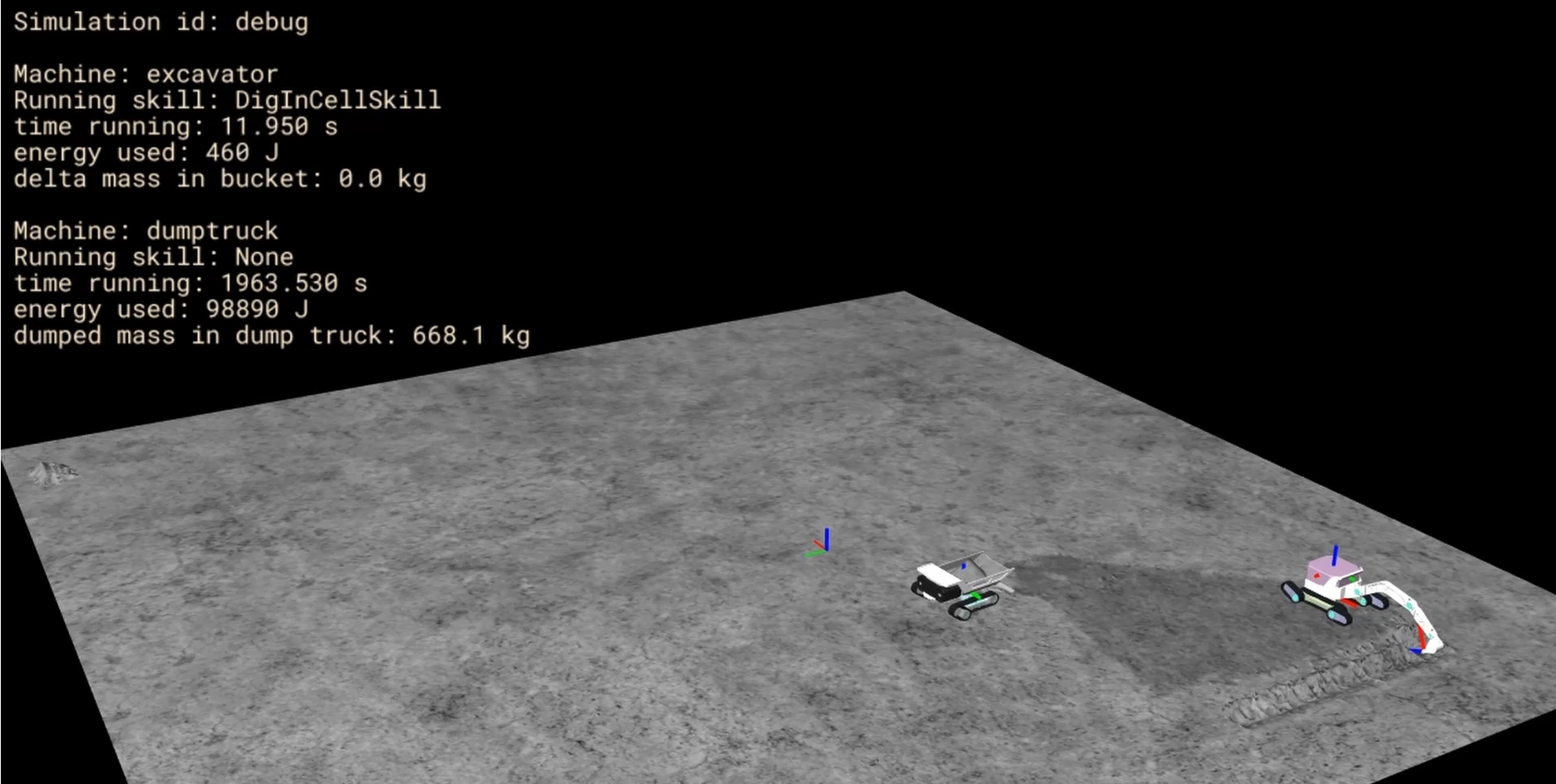}
    \includegraphics[height=0.2\textwidth,trim={0mm 0mm 0mm 0mm},clip]{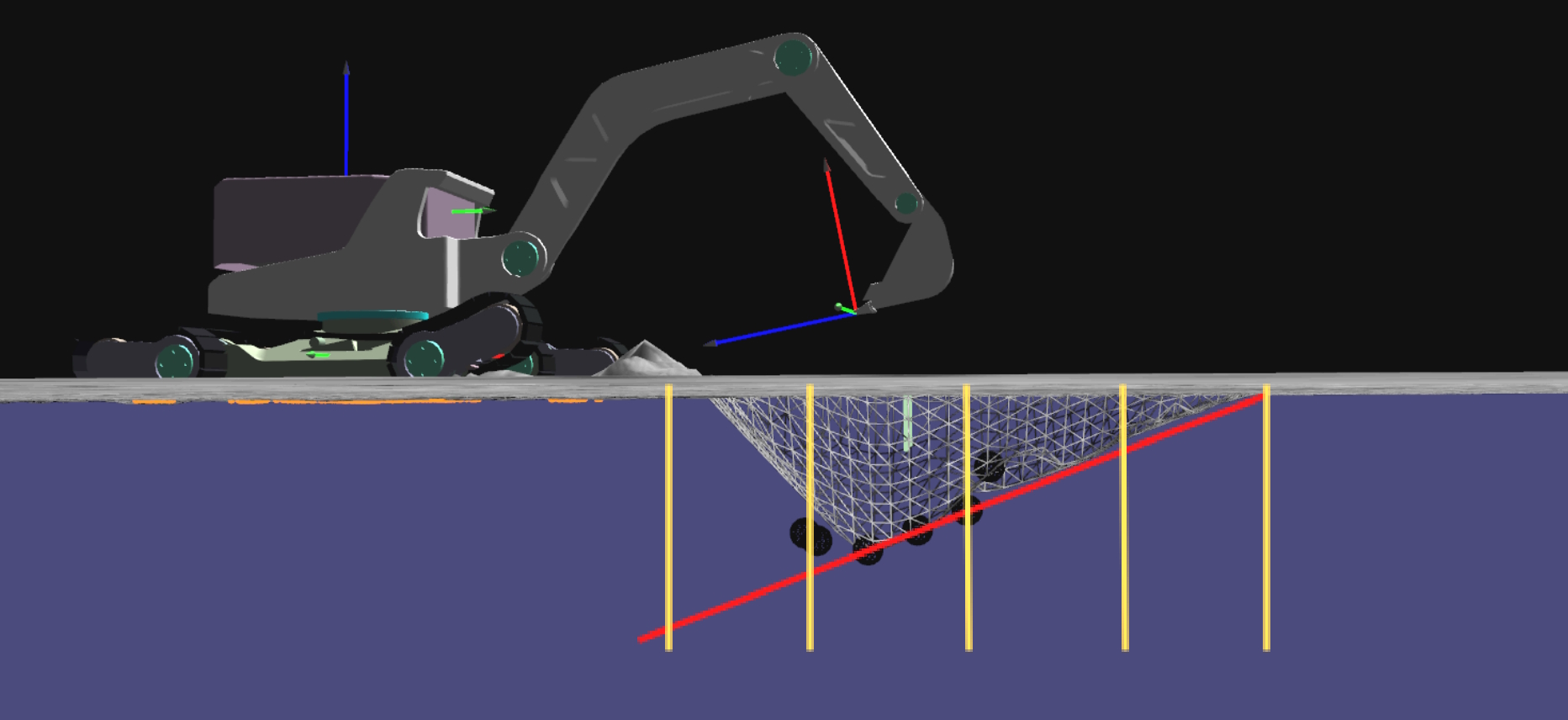}
    \caption{Beginning of excavating the ground construction for habitat modules.
    The lower image shows a cross-section with the target surface indicated in red and the estimated cell boundaries in yellow.}
    \label{fig:beginning-of-ground-construction}
\end{figure}

\begin{figure}[H]
    \centering
    \includegraphics[width=0.65\textwidth,trim={0mm 0mm 0mm 0mm},clip]{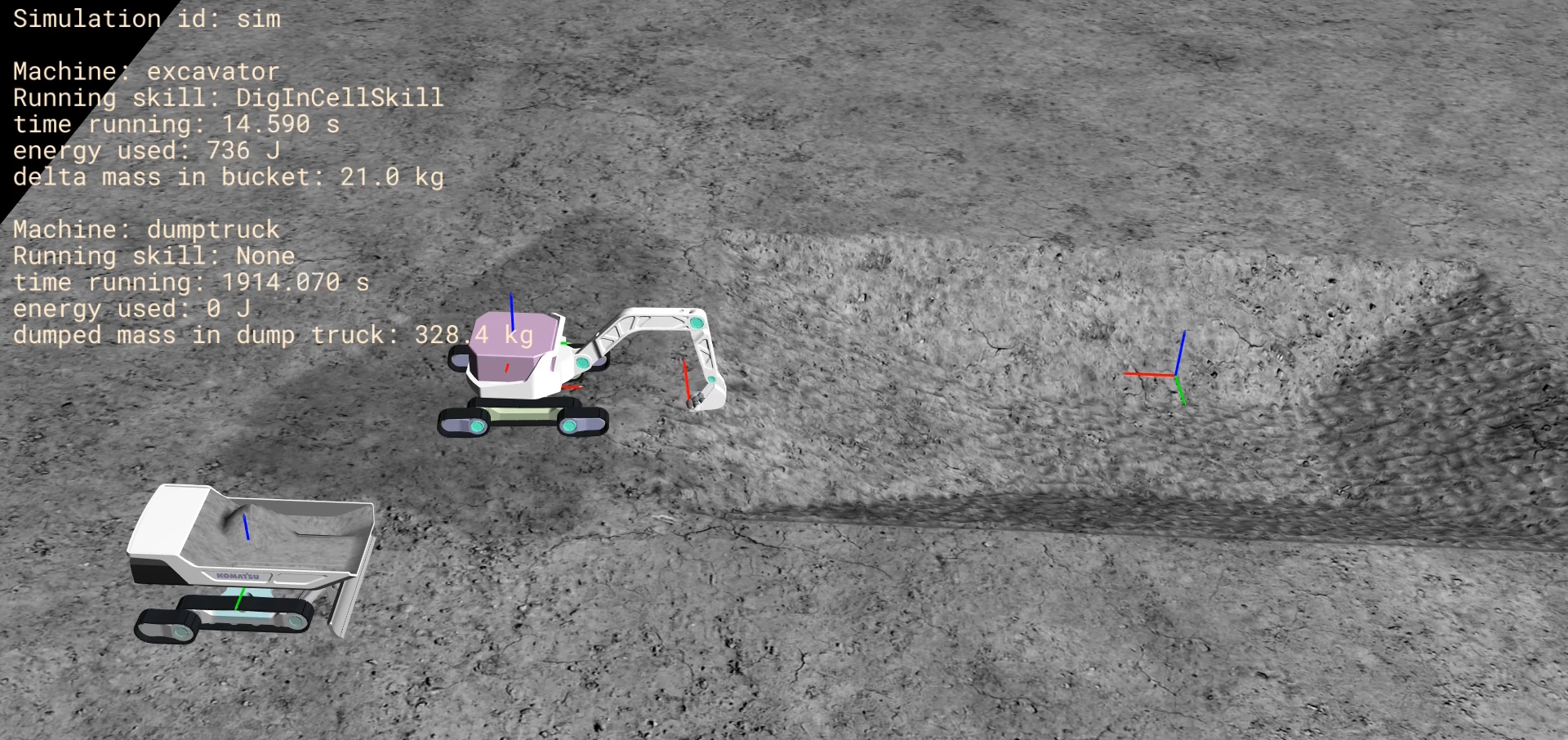}
    \caption{End of excavating the ground construction for habitat modules.}
    \label{fig:end-of-ground-construction}
\end{figure}

The beginning of the scenario is displayed in Fig.~\ref{fig:beginning-of-ground-construction} 
and the end in Fig.~\ref{fig:end-of-ground-construction}.
When a cell is excavated, a new bucket tip trajectory is planned for each dig cycle.
The positions are planned with the excavation depth down from the current terrain surface,
but no deeper than the target surface or maximum digging depth. The target angle for 
the bucket is adapted to the target slope of the terrain.
Excavating the entire structure would require roughly 10,000 work cycles
which would amount to about a week of serial computing time. In our tests, 
we jumped to selected intermediate states and verified the operational 
performance over a sequence of cycles.

After the target 3D profile has been excavated, it is graded.
When “grading” is used, the heights are different during the dig trajectory. If, on the other
hand, the target profile is flat and “trench digging” is used, then a flat dig trajectory is used.
For both cases, the shovel first reaches for a point slightly before and above the trajectory
start, which improves the ability to get to the trajectory start with the specified angle.

During excavation, the terrain will avalanche, both sideways and also from the next row of cells
that are closer to the excavator. Material from cells which have not been excavated yet will
have material removed in the process of getting the current cell to its desired shape.

\begin{figure}[H]
    \centering
    \includegraphics[height=0.25\textwidth,trim={0mm 0mm 0mm 0mm},clip]{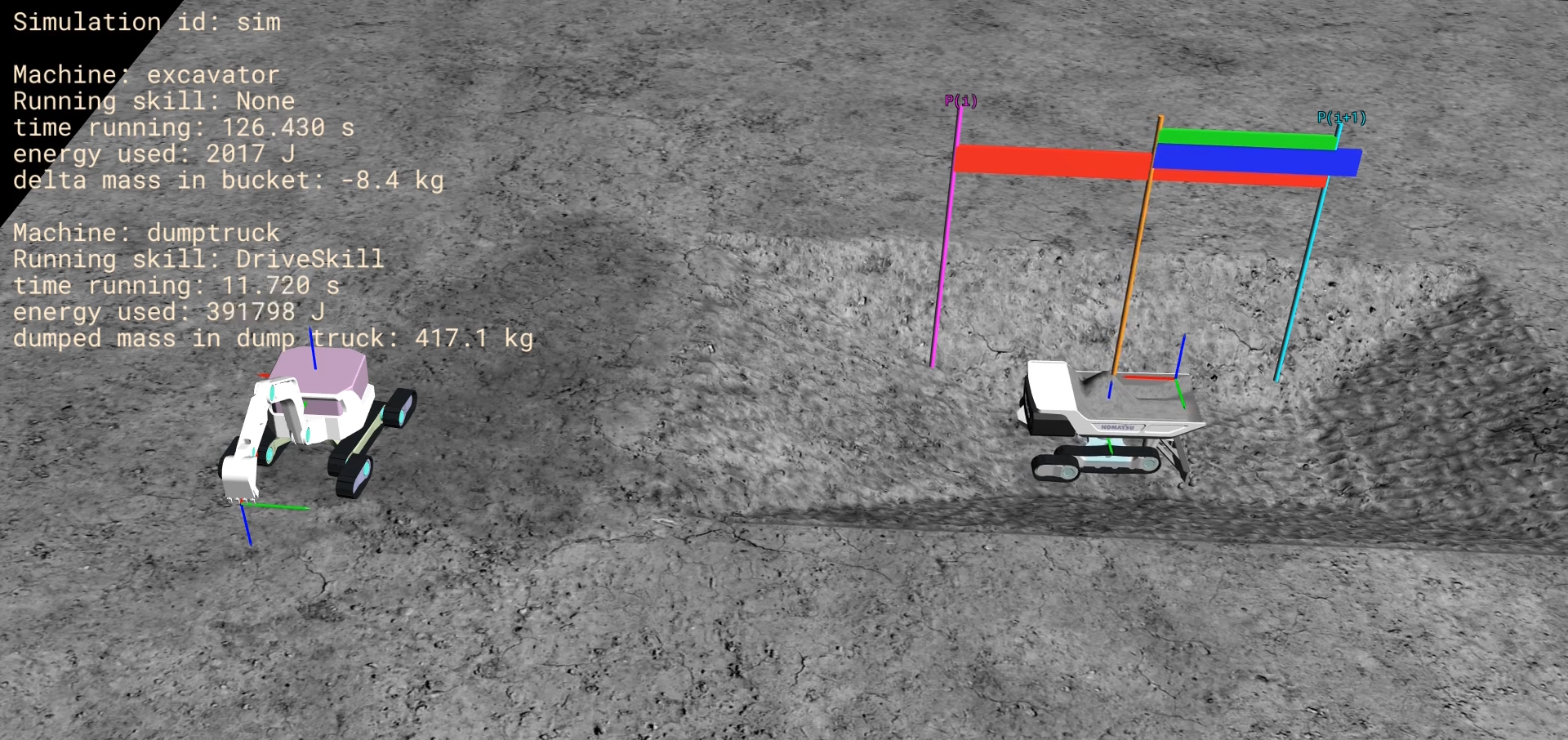}
    \includegraphics[height=0.25\textwidth,trim={0mm 0mm 0mm 0mm},clip]{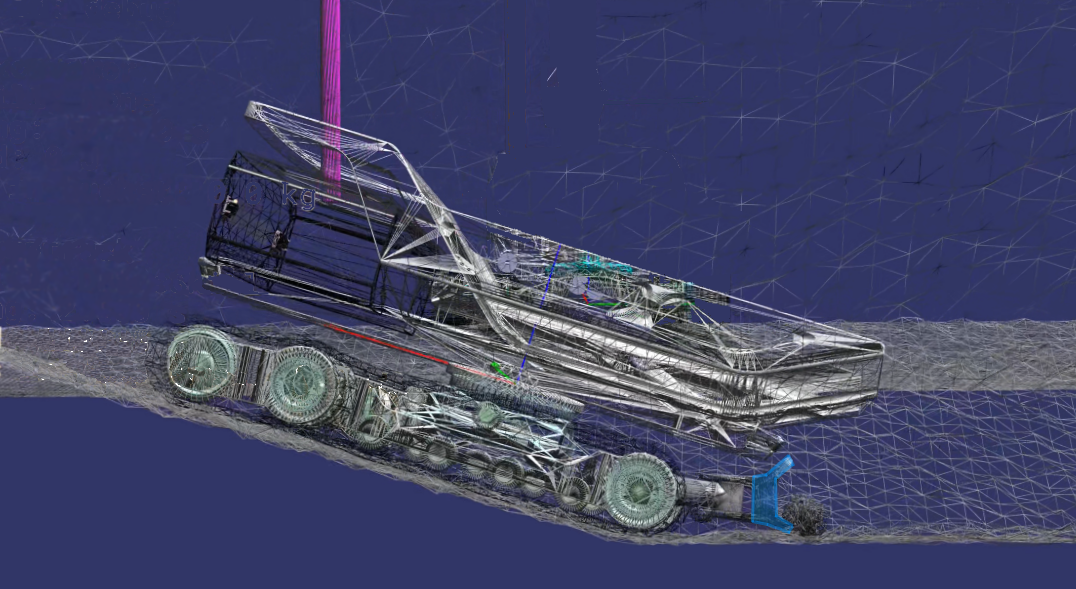}
    \caption{After excavating the ground construction for habitat modules,
    the excavator leaves room for the dump truck to perform final grading of the target surface.
    The lower image is a cross-section of the grading with the actively controlled blade highlighted in blue.}
    \label{fig:grading}
\end{figure}

The dump truck operates as in the previous scenario until the excavator has finished 
grading the target shape. The dump truck is then performing final grading of the 
bottom of the shape using its blade, see Fig.\ref{fig:grading}
The \texttt{leveling/start\_position},
\texttt{leveling/end\_position}, and \texttt{leveling/offset} variables from the last work setting
provided are used to plan out a series of runs across the bottom of the shape. These runs
are then performed with the shovel lowered to level the terrain.
The blade position is dynamically adjusted using a PID controller with parameters provided by the machine manufacturer.

\section{Conclusions}
We have found that the simulation framework serves the purpose of analysing
how a system for autonomous lunar construction performs under different
circumstances, which can include differences in the lunar environment,
machine design and control algorithms, and more.  Future work should include
more elaborate models for sensors and lunar soil, and explore the framework's
scalability for a large number of machines. The normalization procedure for
compensating for numerical damping in the tracks should be experimentally validated
to ensure that the energy computations are reliable. 
Automated generation and improvement
of construction instructions using artificial intelligence is also interesting
to evaluate.

\section{Acknowledgements}
The research was commissioned from Komatsu Ltd, which is commissioned for the Space construction innovation project as part of the Stardust Program in Japan, 
and supported in part by the Swedish National Space Agency through Rymdtillämpningsprogrammet dnr 2024-00310 (AILUR).

\newpage
\printbibliography

\end{document}